\documentclass[runningheads]{llncs}

\usepackage{eccv}

\usepackage{eccvabbrv}

\usepackage{graphicx}
\usepackage{booktabs}
\usepackage{arydshln}
\usepackage{multirow}
\usepackage{colortbl}
\usepackage[misc]{ifsym}
\definecolor{mygray}{gray}{.9}
\usepackage[accsupp]{axessibility}  %

\usepackage{hyperref}

\usepackage{orcidlink}

\newcommand{\dn}{{DepthNet}\xspace}
\newcommand{\kn}{{KeypointNet}\xspace}
\newcommand{\jn}{{JointNet}\xspace}
\newcommand{\rn}{{RotationNet}\xspace}

\makeatletter
\newcommand{\thickhline}{%
    \noalign {\ifnum 0=`}\fi \hrule height 1pt
    \futurelet \reserved@a \@xhline
}

\begin{document}

\renewcommand{\thefootnote}{}
\footnotetext{$^\textrm{\Letter}$Correspondence to: Wentao Zhu <wtzhu@pku.edu.cn> and Yu Qiao <qiaoyu@sjtu.edu.cn>.}

\title{Real-time Holistic Robot Pose Estimation with Unknown States} 

\titlerunning{Holistic Robot Pose}

\author{
Shikun Ban\inst{1}\orcidlink{0009-0007-6330-6057} \and
Juling Fan\inst{1}\orcidlink{0009-0008-5012-8308} \and
Xiaoxuan Ma\inst{1}\orcidlink{0000-0003-0571-2659} \and \\
Wentao Zhu\inst{1, \textrm{\Letter}}\orcidlink{0000-0002-5483-0259} \and
Yu Qiao\inst{5, \textrm{\Letter}}\orcidlink{0000-0001-8258-3868} \and
Yizhou Wang\inst{1,2,3,4}\orcidlink{0000-0001-9888-6409}}

\authorrunning{S.~Ban et al.}

\institute{Center on Frontiers of Computing Studies, School of Computer Science, \\ Peking University \and 
Inst. for Artificial Intelligence, Peking University \and
Nat’l Eng. Research Center of Visual Technology \and
Nat’l Key Lab of General Artificial Intelligence \and
Department of Automation, School of Electronics, Information and Electrical Engineering, Shanghai Jiao Tong University}

\maketitle

\begin{abstract}
  Estimating robot pose from RGB images is a crucial problem in computer vision and robotics. While previous methods have achieved promising performance, most of them presume full knowledge of robot internal states, \eg ground-truth robot joint angles. However, this assumption is not always valid in practical situations. In real-world applications such as multi-robot collaboration or human-robot interaction, the robot joint states might not be shared or could be unreliable. On the other hand, existing approaches that estimate robot pose without joint state priors suffer from heavy computation burdens and thus cannot support real-time applications. This work introduces an efficient framework for real-time robot pose estimation from RGB images without requiring known robot states. Our method estimates camera-to-robot rotation, robot state parameters, keypoint locations, and root depth, employing a neural network module for each task to facilitate learning and sim-to-real transfer. Notably, it achieves inference in a single feed-forward pass without iterative optimization. Our approach offers a 12$\times$ speed increase with state-of-the-art accuracy, enabling real-time holistic robot pose estimation for the first time. Code and models are available at \url{https://github.com/Oliverbansk/Holistic-Robot-Pose-Estimation}.
\end{abstract}

\begin{figure}[t]
  \centering
   \includegraphics[width=\linewidth]{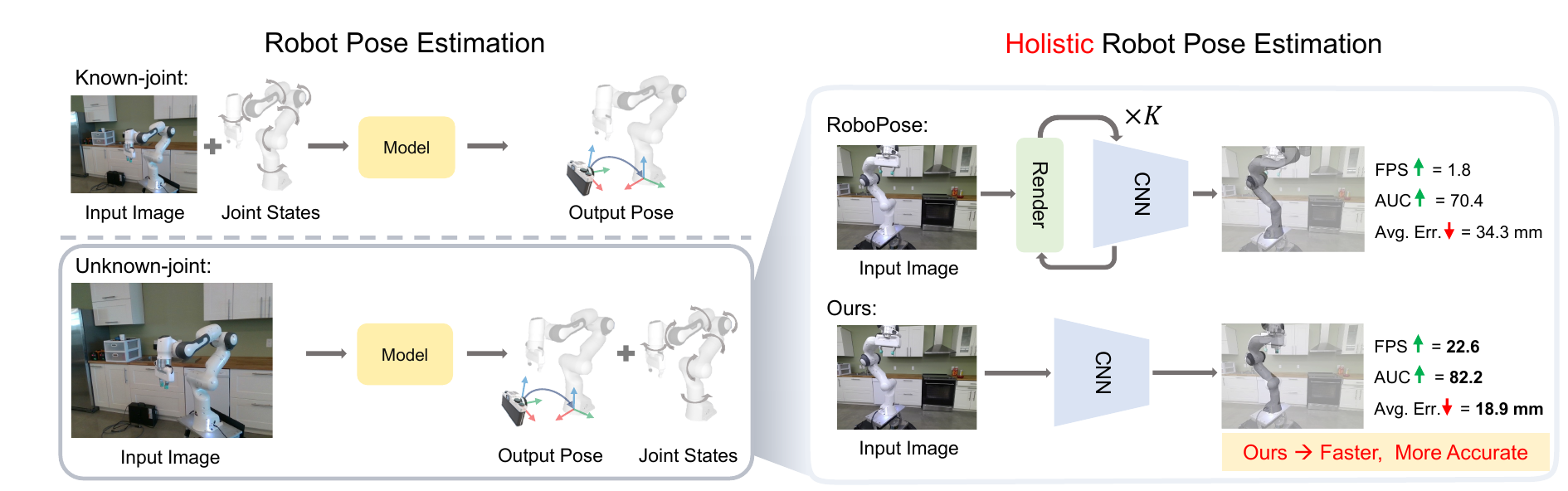}
   \captionof{figure}{The majority of previous robot pose estimation methods assume known robot joint states and focus solely on estimating the camera-to-robot pose, \ie camera-to-robot rotation and translation. In contrast, the holistic robot pose estimation problem requires estimating both joint states and the camera-to-robot pose, given only an RGB image without known joint states. For holistic robot pose estimation, RoboPose~\cite{labbe2021robopose} uses costly test-time optimization (Render-and-Compare) to iteratively refine the predictions. In contrast, our feed-forward method achieves state-of-the-art accuracy with a $12\times$ speed boost.}
   \label{fig:fig1}
    \vspace{-0.2cm}
\end{figure}

\section{Introduction}
\label{sec:intro}

Robot pose estimation is a fundamental task in computer vision and robotics as it not only lays the groundwork for multiple downstream tasks, \eg visually-guided object grasping and manipulation~\cite{zuo2019craves, tian2023robot, LI2023110491grasping}, but also paves the way for a variety of applications, such as autonomous navigation~\cite{bultmann2023external, prusak2008pose, 7846512autonomous, 9988012ICECCME}, human-robot interaction~\cite{yang2021reactive, christen2023learning, 5152690, privacy-preserving, 9255211humanrobot, zhu2023human} and multi-robot collaboration~\cite{li2022self, xun2023crepes, PAPADIMITRIOU2022117052collab,Rizk2019CooperativeHM}. 
Nonetheless, most previous works for monocular robot pose estimation~\cite{lambrecht2019towards,lee2020icra:dream,lu2023markerless,tian2023robot} rely on a known set of robot joint angles as an additional input to alleviate the difficulty of the estimation process. 
Yet, this prerequisite information of the robot state is not always available in practical scenarios:
\begin{itemize}
    \item For instance, low-cost robotic arms without sophisticated sensors or internal monitors can not provide the visual calibration process with readily accessible robot state measurements \cite{deisenroth2011learning, zuo2019craves, s20205919industrial}. 
    \item In the context of multi-robot collaboration, the real-time synchronization of joint state data may be hindered by environmental limitations~\cite{MASEHIAN2017188}, necessitating pose estimation without robot joint state information, and additional monitoring of the joint states.
    \item In domains such as human-robot interaction~\cite{TZAFESTAS2014635, Salvini2021SafetyCE} and competitive robot gaming~\cite{uchibe1996vision, 7989190supervision}, the reliability of the state information from counterpart robots is not always guaranteed, as trust issues may arise in robotic systems~\cite{kok2020trust, kuipers2018can, KIM2020103056}. In this sense, failing to obtain independent estimation of the counterpart robot joint state may lead to potential safety issues.
\end{itemize}

These scenarios further emphasize the urgent need for monocular robot pose estimation with unknown robot joint state.
In this work, we address the challenge of \emph{Holistic Robot Pose Estimation} with unknown internal states for articulated robots. Specifically, given a monocular RGB image, we aim to simultaneously solve for 6D robot pose (3D rotation and translation relative to the camera) and fine-grained robot joint state parameters (including joint angles for revolute joints and displacement for prismatic joints). 

Nonetheless, learning holistic robot pose estimation without joint state information presents significant challenges. 
The primary challenge is the dual ambiguity of unknown joint states and 6D pose. Specifically, estimating the 6D pose of an articulated robot is more complex than for a rigid object due to shape and scale variability from unknown joint states. Conversely, estimating joint states is complicated by the need for 2D-to-3D correspondences and the unknown 6D pose. This intertwined issue requires novel solutions to decouple joint states from 6D pose estimation.
Furthermore, the sophisticated robot morphologies vary from model to model, requiring non-trivial model design. 
In addition, The robot parts are frequently invisible in the monocular image as a result of self-occlusion and truncation, posing more challenges for robust estimation. 
Finally, the domain gap between synthetic training data and real-world images introduces additional challenges for learning-based approaches.

It is worth noting that some previous works~\cite{labbe2021robopose,zuo2019craves} have also addressed holistic estimation with unknown states, tackling some challenges. However, these methods all necessitate iterative optimization at inference, which is computation-intensive and unsuitable for real-time applications.
Motivated by the need for efficient holistic robot pose estimation without knowledge of the robot state, we propose an end-to-end framework that directly estimates camera-to-robot pose and robot joint state parameters, requiring only a single feed-forward pass during inference.

In pursuit of this goal, we factorize the holistic robot pose estimation task into several sub-tasks, including the estimation of camera-to-robot rotation, robot joint states, root depth, and root-relative keypoint locations. We then design corresponding neural network modules for each sub-task. 
Specifically, the standalone root depth estimator disentangles the estimation of scale-variant and scale-invariant variables. The keypoint estimator bridges the gap between the pixel space and the robotic parameter space. Additionally, we utilize self-supervised consistency regularizations which could be performed on the real images without ground truth (GT). 

We demonstrate that our designs of modular feed-forward networks improve model generalization while maintaining high computation efficiency. Consequently, our approach achieves state-of-the-art estimation accuracy with unknown robot states and a $12\times$ speedup compared to the iterative Render-and-Compare (RnC)~\cite{labbe2021robopose} approach, as depicted in~\cref{fig:fig1}.
The key insight behind our approach is that we could get rid of the cumbersome RnC~\cite{labbe2021robopose} and error-prone Perspective-n-Points (P$n$P)~\cite{lee2020icra:dream,lu2023markerless,tian2023robot} schemes to establish 2D-3D correspondences. Instead, we demonstrate that feed-forward networks could well handle the problem with proper problem decomposition and meticulous design.

In summary, our contributions are three-fold:
\begin{enumerate}
    \item We present an efficient end-to-end learning framework for the holistic robot pose estimation problem with unknown robot states.
    \item We propose to factorize and decompose the holistic robot pose estimation problem into several sub-tasks and tackle them with respective modules. 
    \item We achieve state-of-the-art performance on various robot models while delivering a significant $12\times$ speedup, paving the way for real-time applications.
\end{enumerate}

\section{Related Work}
\label{sec:related}

\subsection{Hand-eye calibration} 

The goal of Hand-Eye Calibration (HEC) is to determine the transformation between the position of a robot's end-effector (hand) and its vision system (eye).
This task can be categorized into two types: \emph{eye-in-hand} where the camera is fixed at the robot end-effector~\cite{Morrison_handineye}, and \emph{eye-to-hand} which refers to the configurations where the camera is placed at a distance from the robot~\cite{Feniello2014ProgramSB,DexNet4,Park_handtoeye}. This work focuses on a similar setting with eye-to-hand. Traditional HEC methods involve attaching a fiducial marker to the end effector~\cite{garrido2014automatic,fiala2005artag,olson2011apriltag}, placing it in several known positions, detecting the marker positions in images, and then solving for the camera-to-robot transformation. Despite its broad applicability, the classical HEC pipeline necessitates a complete offline calibration process every time the camera-robot transformation changes, which can be tedious in real-world applications. Consequently, recent works, such as DREAM~\cite{lee2020icra:dream}, propose an online, markerless calibration problem, as discussed below.

\subsection{Image-based robot pose estimation}

Previous works can be divided into two categories based on whether the robot states are known or not. 
Most previous works estimate the 6D camera-to-robot pose with known robot states. For example, DREAM~\cite{lee2020icra:dream} first trains a 2D keypoint estimator and then applies Perspective-$n$-Point (P$n$P)~\cite{lepetit2009epnp} to obtain camera-to-robot pose, showing comparable results with the classical methods. Some work further utilizes segmentation masks and edges to aid pose estimation~\cite{lambrecht2021optimizing,hao2018vision}. Chen~\etal~\cite{chen2023easyhec} apply rendering-based camera pose optimization and consistency-based joint space exploration. SGTAPose~\cite{tian2023robot} leverages robot structure prior and temporal information. Lu~\etal~\cite{lu2023markerless} propose to train on real-world data using foreground segmentation and differentiable rendering in a self-supervised manner.

Recently, several works have shifted to estimating robot pose with unknown joint states. Zuo~\etal~\cite{zuo2019craves} estimate the robot pose and joint angles of a simple 4-DoF robotic arm from a single RGB image by solving a nonlinear nonconvex optimization problem. However, this method is not well-suited for robot arms with higher complexity. Labb\'e~\etal~\cite{labbe2021robopose} propose a render-and-compare approach to iteratively refine the estimation for both known and unknown states, and apply the method to different robot models up to 15-DoF. However, the iterative optimization process limits its applications in real-time scenarios. 
In contrast, our approach can be deployed with a single feed-forward at inference and achieves real-time efficiency.

\begin{figure*}[t]
  \centering
   \includegraphics[width=\linewidth]{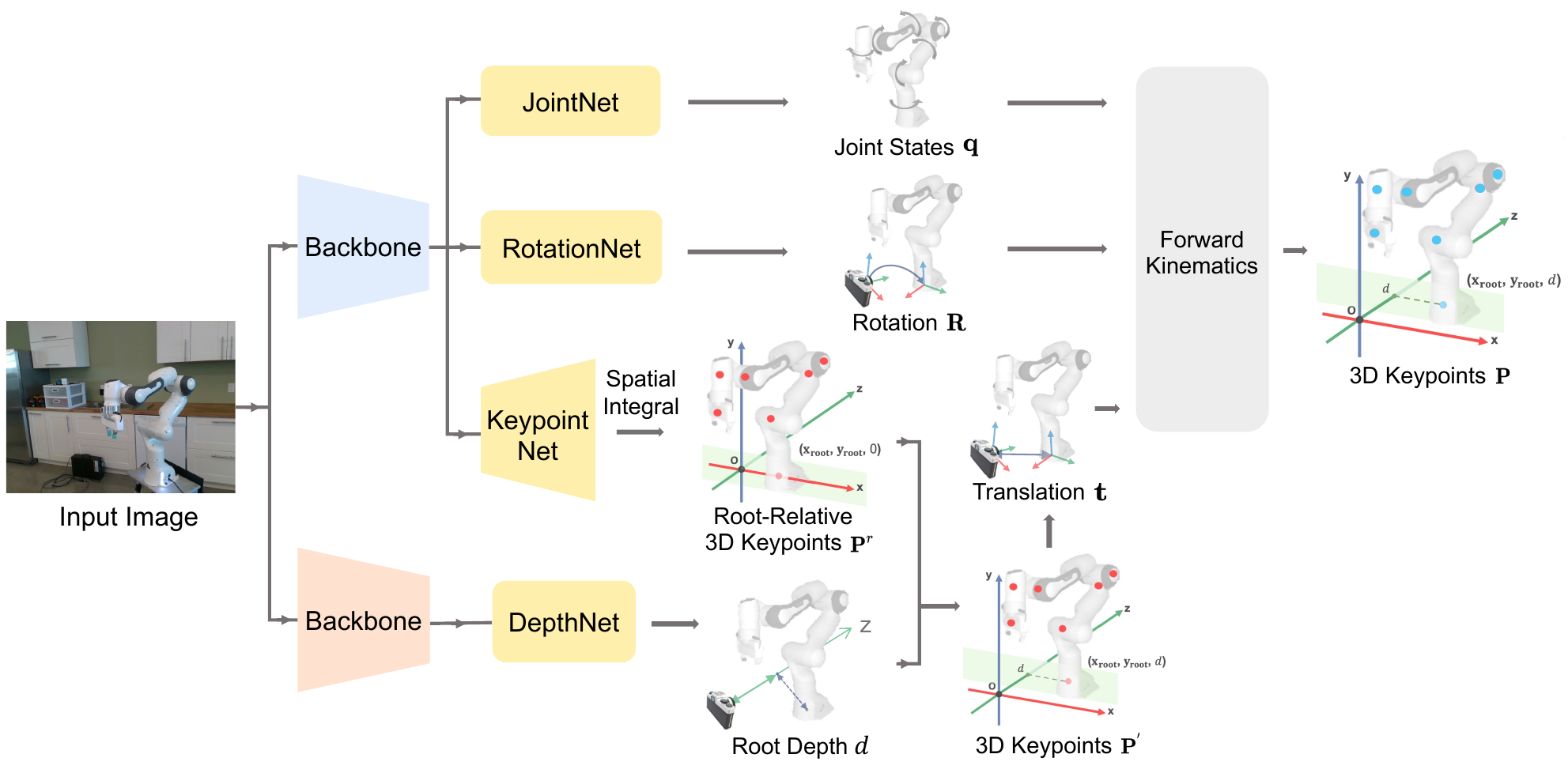}
   \caption{Framework overview. The \jn and the \rn regress joint state parameters $\mathbf{q}$ and camera-to-robot rotation $\mathbf{R}$, respectively. The \kn estimates root-relative 3D keypoint locations $\mathbf{P}^{r}$. The \dn's estimation of root depth $d$ is combined with $\mathbf{P}^{r}$ to acquire absolute 3D keypoint locations $\mathbf{P}^{'}$ and camera-to-robot translation $\mathbf{t}$. Joint state parameters $\mathbf{q}$, rotation $\mathbf{R}$ and translation $\mathbf{t}$ are used to compute 3D keypoint locations $\mathbf{P}$ via forward-kinematics. }
   \label{fig:framework}
\end{figure*}

\section{Method}
\label{sec:method}

\subsection{Overview}
\label{sub:Overview}

Given an RGB image $\mathbf{I}$, we employ a suite of neural network modules, including \dn, \jn, \rn, and \kn. These modules are used to respectively estimate the absolute root depth $d$, joint state parameters $\mathbf{q}$, camera-to-robot rotation $\mathbf{R} \in SO(3)$, and root-relative 3D keypoint locations $\mathbf{P}^{r} \in \mathbb{R}^{N \times 3}$, where $N$ represents the total number of keypoints. The keypoints are defined at the joint articulations, in line with previous works \cite{lee2020icra:dream, labbe2021robopose, lu2023markerless}, and one keypoint is designated as the robot root. The root joint is critical for computing the camera-to-robot rotation $\mathbf{R}$ and translation $\mathbf{t}$, as well as serving as the origin for the forward kinematics (FK) process.

Our modularized design offers several key insights. Firstly, \dn is tailored to learn the scale-variant camera-robot spatial relationship, while the other modules concentrate on the scale-invariant robot features. Secondly, while the joint state parameters $\mathbf{q}$ provide a compact representation of robot morphologies subject to robotic structure constraints, they are highly abstract and thus challenging to directly regress from image features. To address this, we utilize the 3D keypoint locations as an intermediary. On the one hand, they can be effectively estimated with pixel-aligned image features; on the other hand, they are inherently linked to $\mathbf{q}$ through an FK process.

An overview of the proposed framework is illustrated in~\cref{fig:framework}. Specifically, \dn estimates the absolute root depth $d$, which,  combined with \kn's output $\mathbf{P}^{r} \in \mathbb{R}^{N \times 3}$, provides the absolute 3D keypoint locations $\mathbf{P}^{'} \in \mathbb{R}^{N \times 3}$. The root keypoint location in $\mathbf{P}^{'}$ is the camera-to-robot translation $\mathbf{t}$. The estimated joint state parameters $\mathbf{q}$, along with the camera-to-robot rotation $\mathbf{R}$ and translation $\mathbf{t}$, are then used to perform FK, yielding the final 3D keypoint estimations $\mathbf{P} \in \mathbb{R}^{N \times 3}$. Please refer to the appendix for a detailed illustration of the relations of these variables.

In the following, we will first introduce the architecture design of the network modules in \cref{sub:Architecture}, and then explain the training loss design consisting of ground-truth supervision and self-supervision in \cref{sub:Loss}.

\subsection{Architecture}
\label{sub:Architecture}

\subsubsection{DepthNet} 
Monocular object depth estimation suffers from significant ambiguities, as the absolute depth value and camera intrinsics are deeply intertwined~\cite{MING202114depth}. For instance, a robot in a fixed pose relative to the camera may appear larger in photos taken with cameras having larger focal lengths. Scaling both the absolute depth and focal lengths equally results in identical object size and appearance in different images. To address this challenge, we implement \dn following the approach in~\cite{Moon_2019_ICCV_3DMPPE} to disentangle depth from camera intrinsics.
Initially, we use camera intrinsics to calculate a normalized, coarse depth value $d_{c}$:
\begin{equation}
    d_{c} = \sqrt{\frac{f_x \cdot f_y \cdot A_{real}}{A_{bbox}}},
    \label{eq:d_c}
\end{equation}
where \(f_x\) and \(f_y\) represent the focal lengths along the x- and y-axes, \(A_{real}\) is the area of the robot in the x-y plane of the camera frame (in \(mm^2\)), and \(A_{bbox}\) is the area of the robot's bounding box (in \(pixel^2\)). This method of calculating \(d_{c}\) normalizes it in 3D space and provides an approximate depth estimation for subsequent refinement. The primary function of \dn is to predict a correction factor \(\lambda\) based on the input image. We multiply the estimated \(\lambda\) by \(d_{c}\) to obtain the final absolute root depth \(d\):
\begin{equation}
    d = \lambda \cdot d_c.
    \label{eq:dc_to_d}
\end{equation}
With a fixed camera, the relationship between the normalized coarse depth \(d_{c}\) and the robot's size in the image is linear. However, robots can assume poses with the same depth but different sizes in the image, making \(d_{c}\) inadequate for accurate root depth representation. The correction factor \(\lambda\) introduces non-linear flexibility to the depth estimation, addressing this issue by adjusting \(d_{c}\) based on image features.

\subsubsection{JointNet} The goal of \jn is to estimate joint state parameters $\mathbf{q}=(q_{1}, q_{2}, $\ $..., q_J)$, where $J$ denotes the number of robot joints. As this work deals with 1-DoF joints (revolute or prismatic), $J$ equals the robot's DoF.
\jn shares the same feature extractor with \rn and \kn. We extract feature map $\mathbf{F} \in \mathbb{R}^{C \times H \times W}$ (where $C$ denotes the number of channels, and $H \times W$ denotes the spatial resolution) and use global average pooling to obtain $\mathbf{f} \in \mathbb{R}^{C}$. Finally, we employ several MLP blocks to regress the joint state parameters $\mathbf{q}$ from $\mathbf{f}$.

\subsubsection{RotationNet}
The goal of \rn is to estimate the camera-to-robot rotation $\mathbf{R} \in SO(3)$. We employ an MLP to estimate $\mathbf{R}$ from the image feature $\mathbf{f}$. We utilize the 6D continuous rotation representation~\cite{zhou2019continuity} to parameterize $\mathbf{R}$.

\subsubsection{KeypointNet}
The goal of \kn is to estimate root-relative 3D keypoint locations $\mathbf{P}^{r} \in \mathbb{R}^{N \times 3}$ for $N$ robot keypoints. We transform the feature map $\mathbf{F}$ to a 3D heatmap $\mathbf{H} \in \mathbb{R}^{N \times D \times H' \times W'}$ through convolutional layers, where $D \times H' \times W'$ is the spatial resolution discretizing the 3D space. The 3D heatmap encodes per-voxel likelihood of containing each robot keypoint. We derive the root-relative 3D positions of each keypoint $\mathbf{P}^{r}_n$ by performing spatial integral techniques \cite{integral} over the corresponding heatmap $\mathbf{H}_n$ in a differentiable manner:
\begin{equation}
    \mathbf{P}^{r}_n = \sum_{k=1}^{D}\sum_{i=1}^{H'}\sum_{j=1}^{W'}(k, i, j) \cdot \mathbf{H}_n(k, i, j).
\end{equation}
By combining $\mathbf{P}^{r}$ with the estimated root depth $d$, we obtain the regressed 3D keypoint locations $\mathbf{P}^{'}$ and root translation $\textbf{t}$. We further calculate the FK-based 3D keypoint locations $\mathbf{P}$ with differentiable FK parameterized by $\mathbf{q}$, $\mathbf{R}$, and $\mathbf{t}$.

\subsection{Training loss}%
\label{sub:Loss}

\subsubsection{Ground-truth supervision}
\label{sub:GTLoss}

For training on synthetic datasets where we have access to all the ground-truth information, we apply separate supervision for all the sub-tasks to train the corresponding network modules:

\begin{equation}
   \mathcal{L}_{\text{depth}} = \| d - \hat{d} \|_1,
   \label{eq:mse_depth} 
\end{equation}

\begin{equation}
    \mathcal{L}_{\text{joint}} = \| \mathbf{q} - \hat{\mathbf{q}} \|_2,
   \label{eq:mse_q} 
\end{equation}

\begin{equation}
    \mathcal{L}_{\text{rot}} = \| \mathbf{R} - \hat{\mathbf{R}} \|_2,
   \label{eq:mse_rotation} 
\end{equation}

\begin{equation}
    \mathcal{L}_{\text{trans}} = \| \mathbf{t} - \hat{\mathbf{t}} \|_2,
   \label{eq:l2norm_translation} 
\end{equation}

\begin{equation}
    \mathcal{L}_{\text{kpts}} = \| \mathbf{P}-\hat{\mathbf{P}} \|_2 + \| \mathbf{p} -\hat{\mathbf{p}} \|_2,
   \label{eq:l2norm_keypoin_1} 
\end{equation}

\begin{equation}
    \mathcal{L}_{\text{kpts}}^{'} = \| \mathbf{P}^{'}-\hat{\mathbf{P}^{'}} \|_2 + \| \mathbf{p}^{'} -\hat{\mathbf{p}^{'}} \|_2,
   \label{eq:l2norm_keypoint_2} 
\end{equation}
where $d$, $\mathbf{q}$, $\mathbf{R}$, $\mathbf{t}$, $\mathbf{P}$, and $\mathbf{P}^{'}$ are GT root depth, joint states, camera-to-robot rotation and translation, FK-based keypoint locations and regressed keypoint locations, respectively; $\mathbf{p}$ and $\mathbf{p}^{'}$ are 2D projections of $\mathbf{P}$ and $\mathbf{P}^{'}$ using the camera intrinsics; $\hat{}$ denotes model estimations.

We then train all the network modules end-to-end using
\begin{align}
\begin{split}
    \mathcal{L}_{\text{GT}} = \mathcal{L}_{\text{joint}} + \mathcal{L}_{\text{rot}} + \mathcal{L}_{\text{trans}} + \lambda ( \mathcal{L}_{\text{kpts}} + \mathcal{L}_{\text{kpts}}^{'} ),
   \label{eq:overall_loss} 
\end{split}
\end{align}
where we set $\lambda=10.0$ to balance the loss terms.

\subsubsection{Self-supervision}
\label{sub:SelfsupervisionLoss}

Apart from GT supervision on synthetic data, our framework also enables scaling the training pipeline to unlabeled real-world data in a self-supervised manner~\cite{lu2023markerless}. We employ two consistency-based training regularizations to improve model generalization. 

Despite the absence of GT joint states and keypoint locations, we regularize the consistency of the FK-based keypoints $\hat{\mathbf{P}}$ and regressed keypoints $\hat{\mathbf{P}^{'}}$ by:
\begin{equation}
    \mathcal{L}_{\text{kc}} = \| \hat{\mathbf{P}} - \hat{\mathbf{P}^{'}} \|_2.
   \label{eq:l2norm_keypoint_consistency} 
\end{equation}

We introduce this keypoint consistency regularization to effectively reconcile the predictions subject to robot structure constraints $\hat{\mathbf{P}}$ and predictions based on the pixel-aligned feature $\hat{\mathbf{P}^{'}}$.
In addition, similar to~\cite{lu2023markerless}, we also regularize the consistency between robot rendering mask $\mathbf{M}_{\text{render}}$ and foreground segmentation mask $\mathbf{M}_{\text{seg}}$ using their Intersection-over-Union (IoU):

\begin{equation}
    \mathcal{L}_{\text{mc}} = 1 - \frac{\mathcal{S}_{\text{intersection}}}{\mathcal{S}_{\text{union}}},
   \label{eq:iou_mask_consistency} 
\end{equation}
where $\mathcal{S}_{\text{intersection}}$ and $\mathcal{S}_{\text{union}}$ is the area of the intersection and the union of $\mathbf{M}_{\text{render}}$ and $\mathbf{M}_{\text{seg}}$, respectively. $\mathbf{M}_{\text{render}}$ is obtained with differentiable rendering using $\mathbf{q}$, $\mathbf{R}$, and $\mathbf{t}$~\cite{labbe2021robopose}; $\mathbf{M}_{\text{seg}}$ is obtained with an image segmentation model~\cite{lu2023markerless}.

The overall self-supervision training objective is as follows:
\begin{align}
\begin{split}
    \mathcal{L}_{\text{self}} = \mathcal{L}_{\text{kc}} + \lambda_{\text{mc}} \mathcal{L}_{\text{mc}}
   \label{eq:self_supervision_loss} 
\end{split}
\end{align}
where we set $\lambda_{\text{mc}}=1.0$. We apply end-to-end self-supervised training for the sim-to-real domain adaptation stage.

\section{Experiments}
\label{sec:Experiments}

\subsection{Dataset and metrics}
We utilize the benchmark dataset DREAM~\cite{lee2020icra:dream}, for both training and testing in our study. This dataset comprises images from three widely used robotic arms, namely, Franka Emika Panda (Panda), Kuka iiwa7 (Kuka), and Rethink Robotics Baxter (Baxter). The training dataset for each robot consists of approximately 100k synthetic images generated using the domain randomization (DR) technique. These images encompass scenarios with varying camera-to-robot poses and joint states. The testing datasets include both DR-generate synthetic images for each robot and photorealistic synthetic images (denoted as "Photo") for Panda and Kuka. Additionally, real-world image datasets of Panda are provided. Panda-3Cam comprises a total of 17k image frames captured by three different cameras, namely Azure Kinect (AK), XBOX 360 Kinect (XK), and Realsense (RS). This dataset is presented as three sub-datasets, each corresponding to one camera. Another real-world dataset, Panda-ORB, contains 32,315 images captured from 27 viewpoints using a Realsense camera.

We evaluate the accuracy of our estimations using the Average Distance (ADD) metric, which represents the average Euclidean distances (in millimeters) between predicted 3D keypoint locations and their corresponding ground truth locations. Our reported metrics include both the Area Under the Curve (AUC) value and the mean value of ADD. The AUC integrates ADD over different thresholds (with the maximum threshold set at 100mm), while the mean value effectively represents the general prediction accuracy.

\subsection{Implementation details}
We first pre-train \dn on the synthetic training dataset for $100$ epochs, employing a learning rate of $1e-4$ and relying solely on ground truth supervision for depth.
Subsequently, we conduct training for our entire model based on the pretrained \dn for an additional $100$ epochs with a learning rate of $1e-4$ and a decay rate of $0.95$. The Adam optimizer is employed in each stage to optimize the network parameters, with the momentum set to $0.9$. 
As depth influences the global offset of all keypoints, a pre-trained \dn serves as a reasonable starting point for training other networks. 
On the Panda real datasets, we further apply self-supervised training on real-world images to overcome the sim-to-real domain gap with a learning rate of $1e-6$. This fine-tuning process uses only image data and no ground-truth labels.

\subsection{Evaluation and Comparison}

\begin{table*}[h]
    \centering
    \caption{Comparison of AUC $\color{green}\uparrow$ of the ADD distribution curve and mean $\color{red}\downarrow$ of the ADD, on each dataset in the DREAM datasets. Panda 3CAM datasets and Panda ORB are real-world datasets and the rest are synthetic datasets. $\dagger$ denotes using ground-truth joint state parameters. }
    \label{table:main_comparison_auc_mean}
    \setlength{\tabcolsep}{3pt}
    \resizebox{\columnwidth}{!}{
    \begin{tabular}{llccccccccc} 
    \thickhline
    \multirow{2}{*}{Method} & \multirow{2}{*}{Metric} & Baxter & Kuka & Kuka & Panda & Panda & Panda & Panda & Panda & Panda\\
     &  & DR &  DR & Photo & DR & Photo & 3CAM-AK & 3CAM-XK & 3CAM-RS & ORB \\
    \midrule
    $\dagger$ DREAM-F~\cite{lee2020icra:dream} & AUC $\color{green}\uparrow$ & - & - & - & 81.3 & 79.5 & 68.9 & 24.4 & 76.1 & 61.9 \\
    $\dagger$ DREAM-Q~\cite{lee2020icra:dream} & AUC $\color{green}\uparrow$ & 75.5 & - & - & 77.8 & 74.3 & 52.4 & 37.5 & 78.0 & 57.1 \\
    $\dagger$ DREAM-H~\cite{lee2020icra:dream} & AUC $\color{green}\uparrow$ & - & 73.3 & 72.1 & 82.9 & 81.1 & 60.5 & 64.0 & 78.8 & 69.1 \\
    \midrule
    RoboPose~\cite{labbe2021robopose} & AUC $\color{green}\uparrow$ & 32.7 & \textbf{80.2} & 73.2 & \textbf{82.9} & 79.7 & 70.4 & \textbf{77.6} & 74.3 & 70.4 \\
    Ours & AUC $\color{green}\uparrow$ & \textbf{58.8} & 75.1 & \textbf{73.9} & 82.7 & \textbf{82.0} & \textbf{82.2} & 76.0 & \textbf{75.2} & \textbf{75.2} \\
\midrule
    $\dagger$ DREAM-F~\cite{lee2020icra:dream} & Mean $\color{red}\downarrow$ & - & - & - & 552.0 & 785.8 & 11413.1 & 491911.4 & 2077.4 & 95319.1 \\
    $\dagger$ DREAM-Q~\cite{lee2020icra:dream} & Mean $\color{red}\downarrow$ & 59.0 & - & - & 359.0 & 569.0 & 78089.3 & 54178.2 & 27.2 & 64247.6 \\
    $\dagger$ DREAM-H~\cite{lee2020icra:dream} & Mean $\color{red}\downarrow$ & - & 5613.2 & 11099.4 & 325.6 & 518.8 & 56.5 & 7381.6 & 23.6 & 25685.3 \\
    \midrule
    RoboPose~\cite{labbe2021robopose} & Mean $\color{red}\downarrow$ & 87.9 & 34.2 & 89.0 & 21.4 & 32.2 & 34.3 & \textbf{22.3} & 26.0 & 30.1 \\
    Ours & Mean $\color{red}\downarrow$ & \textbf{45.0} & \textbf{28.0} & \textbf{29.0} & \textbf{18.0} & \textbf{18.4} & \textbf{18.9} & 24.0 & \textbf{24.8} & \textbf{24.8}\\
    \thickhline
    \end{tabular}}
\end{table*}

\subsubsection{Accuracy} 
\begin{figure*}[!htbp]
    \centering
   \includegraphics[width=\linewidth]{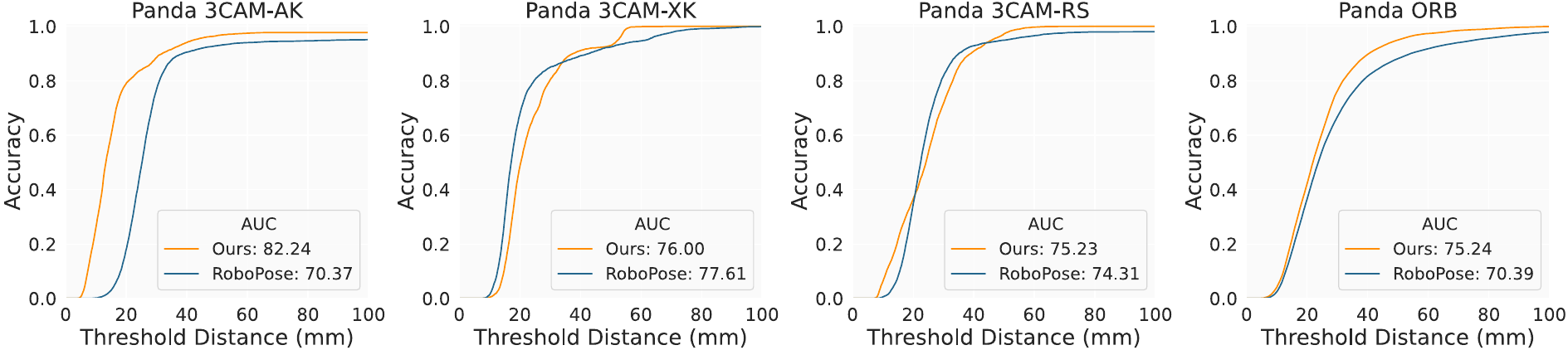}
   \caption{Comparison of ADD distributions on the real-world datasets between our approach and RoboPose~\cite{labbe2021robopose}. The y-axis represents the accuracy of our estimation at different ADD thresholds.}
   \label{fig:add}
\end{figure*}

We first quantitatively evaluate the estimation accuracy of our method and compare it with the state-of-the-art (SOTA) approaches. To the best of our knowledge, among prior works, only RoboPose~\cite{labbe2021robopose} addresses a scenario similar to ours, where both the pose and joint state parameters for an industry-level robot are estimated from a single RGB image. We also present the results of DREAM~\cite{lee2020icra:dream}, a representative feed-forward method using ground-truth joint state parameters, as a reference.
\Cref{table:main_comparison_auc_mean} presents the comparisons of AUC and the mean value of ADD. Our method outperforms SOTA on most datasets, significantly reducing joint localization errors. Specifically, DREAM~\cite{lee2020icra:dream} detects the 2D keypoint locations first, then employs a P$n$P process to solve the 2D-to-3D transformation. However, once the 2D keypoint detector fails due to truncations and occlusions, the P$n$P process produces unreliable numerical solutions. RoboPose~\cite{labbe2021robopose} utilizes iterative render-and-compare, which is prone to falling into local optima on complex robot models with symmetric structures (\eg, Baxter). In contrast, our approach directly regresses both the robot pose and joint state parameters, enabling end-to-end optimization from multi-facet supervision signals. We also visualize the complete ADD distribution. As shown in~\cref{fig:add}, our approach achieves higher accuracy across most ADD thresholds. 
Notably, our approach attains near $100\%$ accuracy with a $60$mm threshold, indicating more reliable estimation results on challenging cases with fewer outliers.

\begin{figure*}[t]
  \centering
   \includegraphics[width=0.9\linewidth]{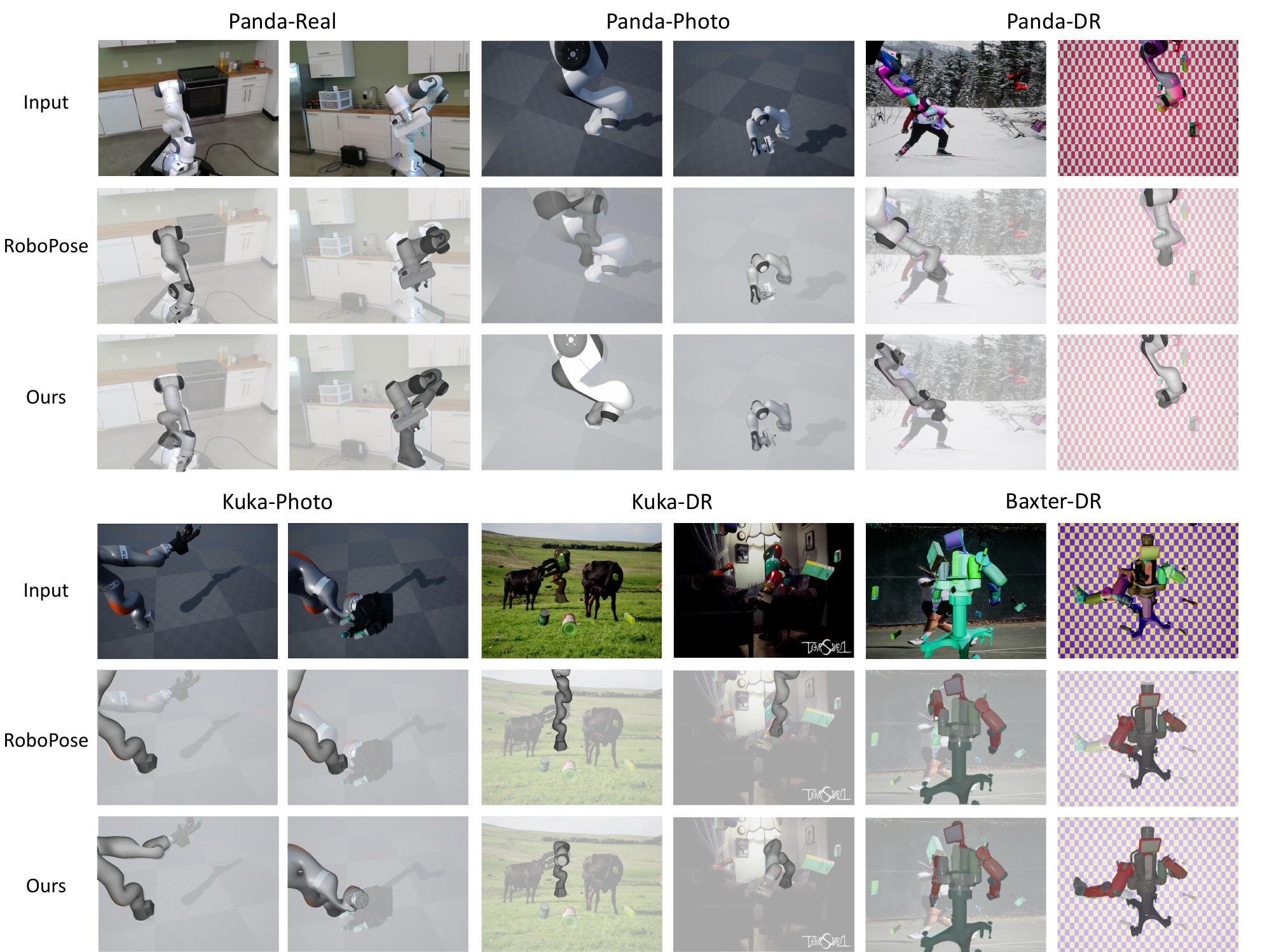}
   \caption{Qualitative comparison between our method and RoboPose~\cite{labbe2021robopose} on both real and synthetic datasets.}
   \label{fig:qualitative}
   \vskip -5pt
\end{figure*}

Finally, we visualize the predictions of our approach and compare them with SOTA to better understand the performance improvement. \Cref{fig:qualitative} illustrates our approach's ability to produce high-quality estimates across different robot morphologies.  
Particularly, our approach demonstrates superior performance in highly challenging cases, including self-occlusions, truncations, and extreme lighting conditions. 
Additionally, we compare the model performance qualitatively on in-the-wild lab images and quantitatively under various levels of truncation severity, illustrating the superior generalization ability and robustness of our approach. Please refer to the appendix for more detailed results.

\subsubsection{Efficiency}
Furthermore, we benchmark the computation efficiency of our method and compare it with previous works. We measure the average inference time, frame per second (FPS), and Floating Point Operations (FLOPs). For a fair comparison, we benchmark all methods on a Linux machine with CPU Intel(R) Xeon(R) Gold 6240 CPU @ 2.60GHz and GPU NVIDIA Tesla V100S. Batch size is set to be 1 for all methods.
\Cref{table:speed_comparison} shows that our method exhibits a remarkable advantage in terms of inference
speed and computation loads, enabling real-time inference while achieving state-of-the-art accuracy. This advantage stems from the fact that our method involves only a single feed-forward pass without costly iterative optimization during inference. We demonstrate that our approach serves as an efficient and effective solution for real-time holistic robot pose estimation, which holds great potential in real-world applications.

\begin{table}[t]
    \centering
    \caption{Comparison of computation efficiency on the Panda 3CAM-AK dataset. $\dagger$ denotes using ground-truth joint state parameters.}
     \label{table:speed_comparison}
    \setlength{\tabcolsep}{4pt}
    \resizebox{0.6\linewidth}{!}{
    \begin{tabular}{lcccc}
    \toprule
    Method & AUC $\color{green}\uparrow$ & Time (ms) $\color{red}\downarrow$  & FPS $\color{green}\uparrow$  & FLOPS (G) $\color{red}\downarrow$ \\ %
    \midrule
    $\dagger$ DREAM-F \cite{lee2020icra:dream} & 68.9 & 66.0 & 15.1 & 90.5 \\ %
    RoboPose \cite{labbe2021robopose} & 70.4 & 570.8 & 1.8 & 57.3 \\ %
    Ours & $\textbf{82.2}$ & $\textbf{44.3}$ & $\textbf{22.6}$ & $\textbf{25.2}$ \\ %
    \bottomrule
    \end{tabular}}
\end{table}

\subsection{Comparison under known joints states} 
Although the primary focus of this work is holistic robot pose estimation with unknown states, our approach can also be easily adapted to the known joint states setting. We conduct additional experiments on four real datasets for robot pose estimation with ground-truth joint states available. We simply replace the predicted joint state with the ground-truth joint state during the self-supervised training process. As shown in~\cref{table:comparison_known_joint}, our approach also achieves comparable performance with previous SOTA under this well-explored setting.

\begin{table*}[t]
    \centering
    \caption{Comparison of AUC $\color{green}\uparrow$ of the ADD distribution curve and Mean $\color{red}\downarrow$ of the ADD, on the real-world datasets of DREAM, with all methods using \textbf{known ground-truth joint state parameters}.}
    \label{table:comparison_known_joint}
    \setlength{\tabcolsep}{6pt}
    \resizebox{0.95\linewidth}{!}{
    \begin{tabular}{llcccc}
    \toprule
    Method & Metric & Panda 3CAM-AK & Panda 3CAM-XK & Panda 3CAM-RS & Panda ORB \\
    \midrule
    DREAM-F \cite{lee2020icra:dream} & AUC $\color{green}\uparrow$ & 68.9 & 24.4 & 76.1 & 61.9 \\
    DREAM-Q \cite{lee2020icra:dream} & AUC $\color{green}\uparrow$ & 52.4 & 37.5 & 78.0 & 57.1 \\
    DREAM-H \cite{lee2020icra:dream} & AUC $\color{green}\uparrow$ & 60.5 & 64.0 & 78.8 & 69.1 \\
    RoboPose \cite{labbe2021robopose} & AUC $\color{green}\uparrow$ & 76.5 & \textbf{86.0} & 76.9 & 80.5 \\
    CtRNet \cite{lu2023markerless} & AUC $\color{green}\uparrow$ & 89.9 & 79.5 & 90.8 & 85.3 \\
    SGTAPose \cite{tian2023robot} & AUC $\color{green}\uparrow$  & 67.8 & 2.1 & 87.6 & 72.3 \\
    Ours  & AUC $\color{green}\uparrow$  & \textbf{90.2} & 81.2 & \textbf{91.9} & \textbf{87.6} \\
    \midrule
    DREAM-F~\cite{lee2020icra:dream} & Mean $\color{red}\downarrow$ & 11413.1 & 491911.4 & 2077.4 & 95319.1 \\
    DREAM-Q~\cite{lee2020icra:dream} & Mean $\color{red}\downarrow$ & 78089.3 & 54178.2 & 27.2 & 64247.6 \\
    DREAM-H~\cite{lee2020icra:dream} & Mean $\color{red}\downarrow$ & 56.5 & 7381.6 & 23.6 & 25685.3 \\
    RoboPose~\cite{labbe2021robopose} & Mean $\color{red}\downarrow$ & 24.2 & \textbf{14.0} & 23.1 & 19.4 \\
    CtRNet~\cite{lu2023markerless} & Mean $\color{red}\downarrow$ & 13.0 & 32.0 & \textbf{10.0} & 21.0 \\
    SGTAPose~\cite{tian2023robot} & Mean $\color{red}\downarrow$ & 35.7 & 157.7 & 12.8 & 35.0 \\
    Ours  & Mean $\color{red}\downarrow$ & \textbf{12.9} & 19.9 & 11.0 & \textbf{15.9} \\
    \bottomrule
    \end{tabular}}
\end{table*}

\subsection{Ablation study}

\begin{table}[htbp]
    \centering
    \begin{minipage}[t]{0.47\textwidth}
        \centering
        \caption{Ablation studies on the effect of separating \dn image feature extractor.}
        \small
        \label{table:ablation_separate_depthnet}
        \setlength{\tabcolsep}{1mm}
        \resizebox{\linewidth}{!}{
        \begin{tabular}{lccc}
        \toprule
        Dataset & Separate & AUC $\color{green}\uparrow$ & Mean (mm)$\color{red}\downarrow$ \\
        \midrule %
        Panda DR (a) & $-$ & 79.98 & 21.9 \\
        Panda DR (b) & $+$ & \textbf{82.68} & \textbf{18.0} \\ 
        \midrule %
        Panda Photo (a) & $-$ & 79.66 & 22.0 \\
        Panda Photo (b) & $+$ & \textbf{81.98} & \textbf{18.4} \\ 
        \bottomrule
        \end{tabular}}
    \end{minipage}
    \hfill
    \begin{minipage}[t]{0.47\textwidth}
        \centering
        \caption{Ablation studies on the effect of pre-training \dn in advance of jointly training.}
        \label{table:ablation_pretrain_depthnet}
        \setlength{\tabcolsep}{1mm}
        \resizebox{\linewidth}{!}{
        \begin{tabular}{lccc}
        \toprule
        Dataset & Pretrain & AUC $\color{green}\uparrow$ & Mean (mm)$\color{red}\downarrow$ \\
        \midrule %
        Panda DR (a) & $-$ & 74.29 & 26.8 \\
        Panda DR (b) & $+$ & \textbf{82.68} & \textbf{18.0} \\ 
        \midrule %
        Panda Photo (a) & $-$ & 73.34 & 27.9 \\
        Panda Photo (b) & $+$ & \textbf{81.98} & \textbf{18.4} \\ 
        \bottomrule
        \end{tabular}}
    \end{minipage}
\end{table}

\subsubsection{Ablation on network modules}
We first conduct an ablation study on the proposed network modules to investigate their effectiveness. Since \jn and \rn are the major regression targets of our task, we focus on the ablation of two assistive network modules, \dn and \kn. We conduct the experiments on the synthetic datasets, utilizing only ground-truth supervision to simplify the process. 

Firstly, we delve into the implementation of \dn. In \cref{table:ablation_separate_depthnet}, we evaluate the design choice of whether to share the backbone of \dn with other modules or not. The results indicate that separating the \dn backbone consistently improves performance. This improvement could be explained by the fact that \dn involves learning scale-variant features, while the other network modules all focus on learning scale-invariant features. We also investigate whether to pre-train the \dn or jointly train it with other modules from scratch. \Cref{table:ablation_pretrain_depthnet} suggests that a pre-training stage leads to considerable performance improvement. This improvement likely occurs because a well-trained \dn reliably aids other modules in better training, as it determines the global translation of all joints.
Additionally, we select the keypoint closest to the robot center as the root keypoint requiring depth prediction by \dn. Please refer to the S2.3 section of the appendix for rationale behind this selection.

Moreover, we conduct ablation experiments to investigate the impact of \kn on other network modules, namely \jn and \rn. We utilize Mean Joint Error, which represents the mean error of the joint state parameter predictions, and the Mean Rotation Angle Error \cite{labbe2021robopose}, which represents the mean error of the Euler angle predictions ($\theta_x^{pred}$, $\theta_y^{pred}$, $\theta_z^{pred}$) for the camera-to-robot rotation, to quantify the accuracy of the \jn and \rn predictions, respectively. It is important to note that a Panda robot consists of seven revolute joints with angle parameters (representing one-DoF rotations) and one prismatic joint with a length parameter (representing one-DoF translation). Therefore, we calculate and present the Mean Joint Error separately for the revolute and prismatic joints of Panda.
We present the detailed ablation results in the S2.1 section of the appendix, in which \Cref{table:ablation_keypointnet_on_jointnet,table:ablation_keypointnet_on_rotationnet} show that the presence of \kn significantly improves the prediction accuracy of \jn and \rn. By sharing a feature extractor with \jn and \rn, \kn guides the network towards learning pixel-aligned robot visual features, which further assists in estimating robot pose and joint states.

Finally, we examine the overall effect of \dn and \kn. To replace \dn, we implement a straightforward alternative, directly regressing the camera-to-robot translation $\mathbf{t}$ from image features $\mathbf{f}$ using an MLP, similar to \rn. 
\Cref{fig:ablation_net} illustrates that both modules play an important role in prediction accuracy. The \dn notably enhances accuracy by improving root depth estimation, a global offset that affects the precision of all joints. Meanwhile, \kn further benefits the estimation results. 

The contribution of these modules to the overall performance validates our design. \dn disentangles depth from camera intrinsics, facilitating a learning-based approach to depth estimation. This method exhibits greater expressive capability compared to the analytical, non-parameterized P$n$P methods employed in previous work\cite{lee2020icra:dream, lu2023markerless, tian2023robot}. On the other hand, \kn enhances both \jn and \rn by steering the shared backbone towards learning more pixel-aligned features, thus complementing the learning objectives defined in the robot kinematic space.

\subsubsection{Ablation on self-supervision}
\begin{figure}[htbp]
    \centering
    \begin{subfigure}[b]{0.47\textwidth}
        \centering
        \includegraphics[width=\textwidth]{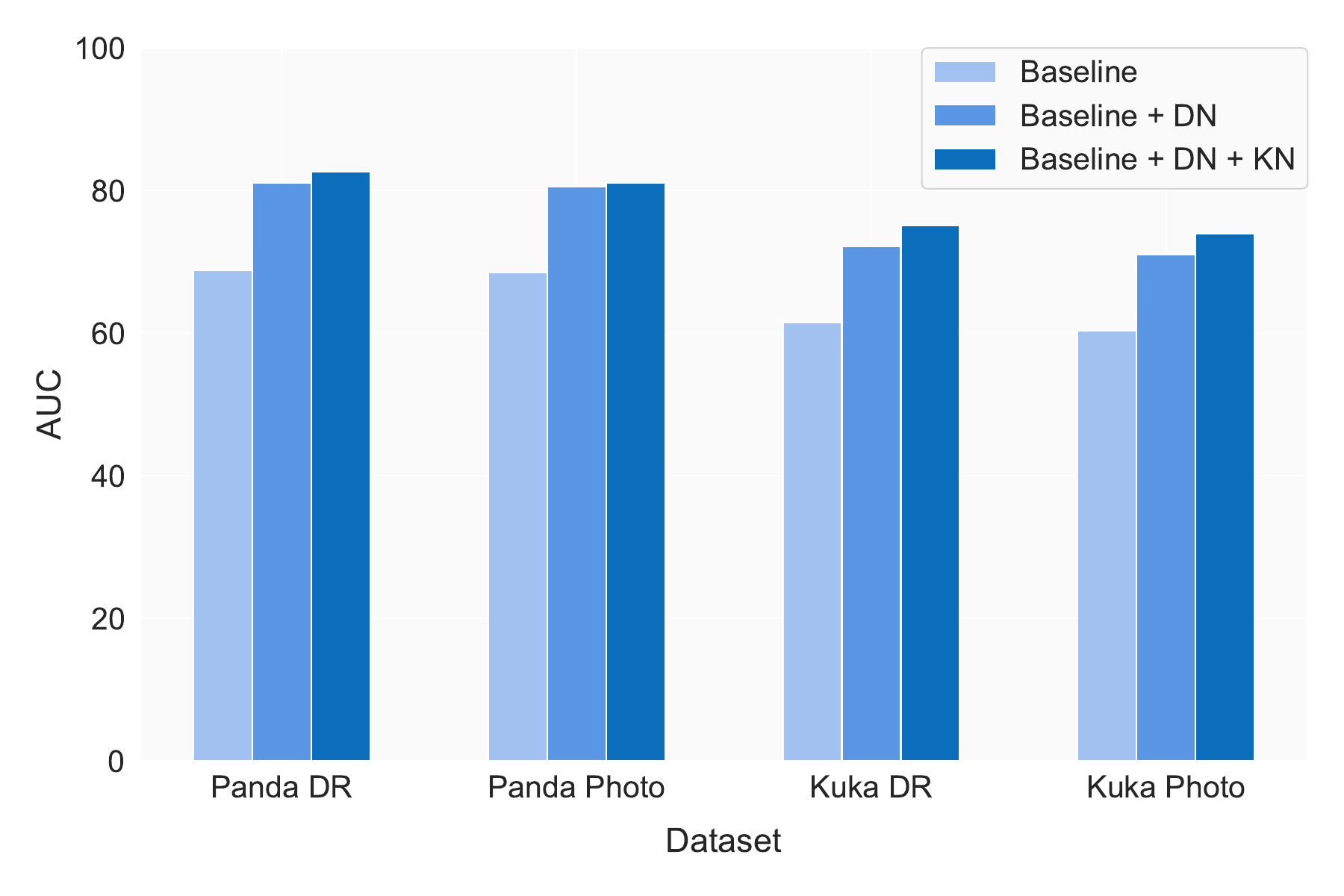}
        \caption{Ablation study on network modules. `Baseline' represents direct regression, `DN' and `KN' refer to \dn and \kn, respectively.}
        \label{fig:ablation_net}
    \end{subfigure}
    \hfill
    \begin{subfigure}[b]{0.47\textwidth}
        \centering
        \includegraphics[width=\textwidth]{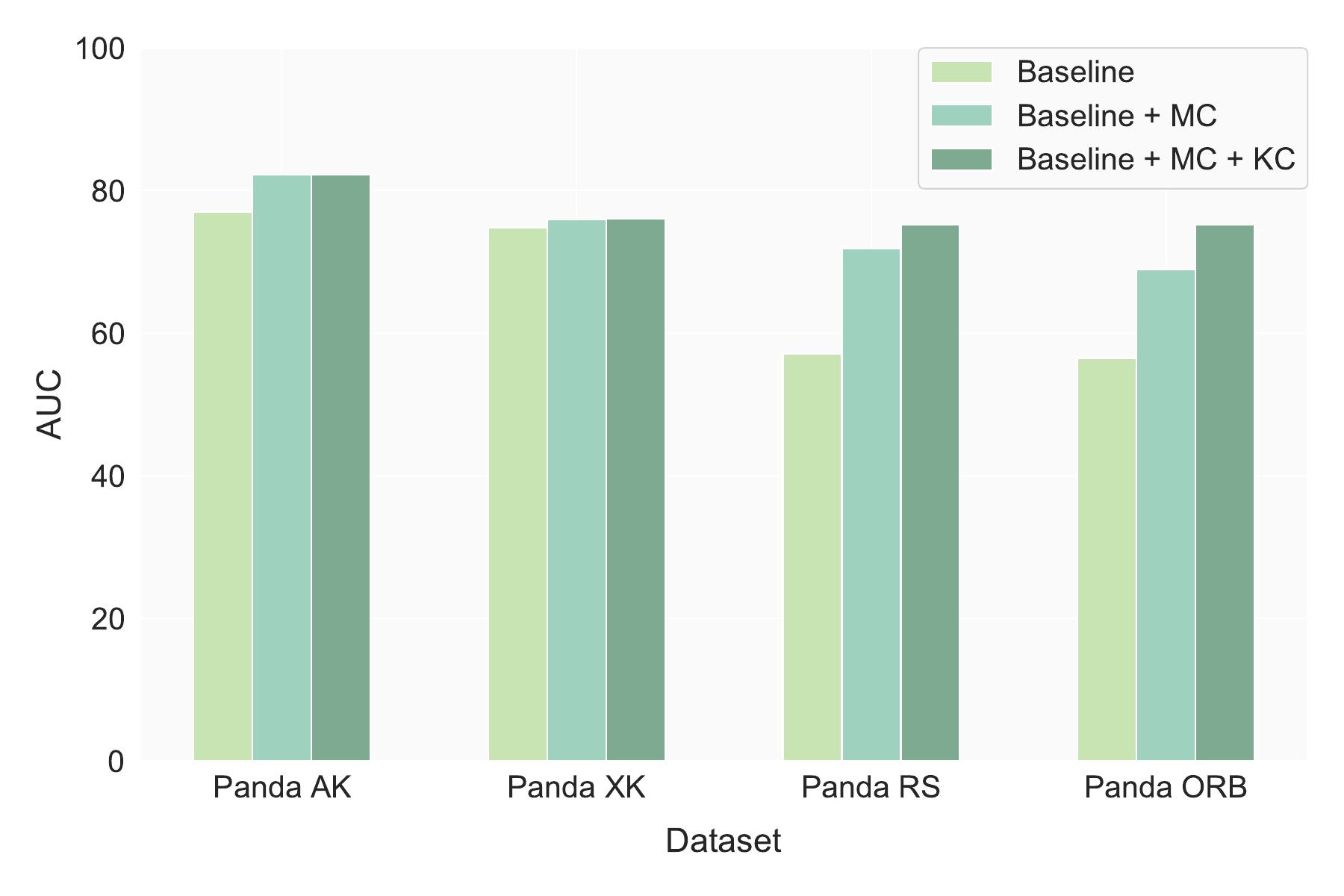}
        \caption{Ablation study on self-supervision. `Baseline' represents no self-supervision, `MC' and `KC' refer to mask and keypoint consistency regularization, respectively.}
        \label{fig:ablation_self}
    \end{subfigure}
    \caption{Ablation studies on network modules and self-supervision.}
    \label{fig:ablation_overall}
\end{figure}

We also perform an ablation study of our self-supervised learning objectives. We conduct the experiments on the Panda real datasets. \Cref{fig:ablation_self} demonstrates that both the keypoint consistency regularization and the mask consistency regularization contribute to the overall performance. Specifically, the mask consistency regularization provides pixel-wise supervision signal with the help of differentiable rendering, while the keypoint consistency regularization unifies the predictions in the pixel space and that in the robot kinematic space.
We suppose that pixel-aligned 2D tasks, \eg keypoint detection and mask segmentation, generally generalize better, and could therefore improve the 3D estimation quality through consistency regularization, further improving the overall performance. As a result, our multi-stage training scheme yields favorable convergence ability. 
Please refer to the S4 section of the appendix for the error analysis of each module across each training stage.

\section{Conclusion}%
\label{sec:Conclusion}

This work presents a novel end-to-end framework for real-time holistic robot pose estimation without prior knowledge of internal robot states, providing substantial assistance in real-world scenarios where the joint state of the estimation target is unavailable or unreliable.
By decomposing this complex task into specific sub-tasks and tailoring neural network modules for each, the method effectively addresses the challenges while retaining high computation efficiency. Our approach, which circumvents iterative optimization, significantly accelerates the estimation process and achieves state-of-the-art accuracy. This advancement holds promise for enhancing real-world robotics applications, offering a practical solution for real-time robot pose estimation and joint state monitoring.

\section*{Acknowledgments}
This work is supported by National Science and Technology
Major Project (2022ZD0114904). We thank Hai Ci for the valuable discussions during the early stages of this work. We thank Haoran Lu and Ruihai Wu for their assistance with real-world image captures.

\bibliographystyle{splncs04}
\bibliography{main}

\begin{thebibliography}{10}
\providecommand{\url}[1]{\texttt{#1}}
\providecommand{\urlprefix}{URL }
\providecommand{\doi}[1]{https://doi.org/#1}

\bibitem{bultmann2023external}
Bultmann, S., Memmesheimer, R., Behnke, S.: External camera-based mobile robot pose estimation for collaborative perception with smart edge sensors. In: ICRA (2023)

\bibitem{chen2023easyhec}
Chen, L., Qin, Y., Zhou, X., Su, H.: {E}asy{H}e{C}: {A}ccurate and {A}utomatic {H}and-eye {C}alibration via {D}ifferentiable {R}endering and {S}pace {E}xploration. In: IEEE Robotics and Automation Letters (2023)

\bibitem{christen2023learning}
Christen, S., Yang, W., P{\'e}rez-D’Arpino, C., Hilliges, O., Fox, D., Chao, Y.W.: Learning human-to-robot handovers from point clouds. In: Proceedings of the IEEE/CVF Conference on Computer Vision and Pattern Recognition. pp. 9654--9664 (2023)

\bibitem{deisenroth2011learning}
Deisenroth, M.P., Rasmussen, C.E., Fox, D.: Learning to control a low-cost manipulator using data-efficient reinforcement learning. Robotics: Science and Systems VII  \textbf{7},  57--64 (2011)

\bibitem{Feniello2014ProgramSB}
Feniello, A., Dang, H.N., Birchfield, S.: Program synthesis by examples for object repositioning tasks. 2014 IEEE/RSJ International Conference on Intelligent Robots and Systems pp. 4428--4435 (2014)

\bibitem{fiala2005artag}
Fiala, M.: Artag, a fiducial marker system using digital techniques. In: CVPR (2005)

\bibitem{garrido2014automatic}
Garrido-Jurado, S., Mu{\~n}oz-Salinas, R., Madrid-Cuevas, F.J., Mar{\'\i}n-Jim{\'e}nez, M.J.: Automatic generation and detection of highly reliable fiducial markers under occlusion. Pattern Recognition  (2014)

\bibitem{hao2018vision}
Hao, R., {\"O}zg{\"u}ner, O., {\c{C}}avu{\c{s}}o{\u{g}}lu, M.C.: Vision-based surgical tool pose estimation for the da vinci{\textregistered} robotic surgical system. In: 2018 IEEE/RSJ international conference on intelligent robots and systems (IROS). pp. 1298--1305. IEEE (2018)

\bibitem{he2016deep}
He, K., Zhang, X., Ren, S., Sun, J.: Deep residual learning for image recognition. In: CVPR (2016)

\bibitem{s20205919industrial}
Icli, C., Stepanenko, O., Bonev, I.: New method and portable measurement device for the calibration of industrial robots. Sensors  \textbf{20}(20) (2020)

\bibitem{KIM2020103056}
Kim, W., Kim, N., Lyons, J.B., Nam, C.S.: Factors affecting trust in high-vulnerability human-robot interaction contexts: A structural equation modelling approach. Applied Ergonomics  \textbf{85},  103056 (2020)

\bibitem{kok2020trust}
Kok, B.C., Soh, H.: Trust in robots: Challenges and opportunities. Current Robotics Reports  \textbf{1},  297--309 (2020)

\bibitem{7846512autonomous}
Kothari, N., Gupta, M., Vachhani, L., Arya, H.: Pose estimation for an autonomous vehicle using monocular vision. In: 2017 Indian Control Conference (ICC). pp. 424--431 (2017)

\bibitem{kuipers2018can}
Kuipers, B.: How can we trust a robot? Communications of the ACM  \textbf{61}(3),  86--95 (2018)

\bibitem{labbe2021robopose}
{Labb\'e}, Y., {Carpentier}, J., {Aubry}, M., {Sivic}, J.: Single-view robot pose and joint angle estimation via render \& compare. In: Proceedings of the Conference on Computer Vision and Pattern Recognition (CVPR) (2021)

\bibitem{lambrecht2021optimizing}
Lambrecht, J., Grosenick, P., Meusel, M.: Optimizing keypoint-based single-shot camera-to-robot pose estimation through shape segmentation. In: 2021 IEEE International Conference on Robotics and Automation (ICRA). pp. 13843--13849. IEEE (2021)

\bibitem{lambrecht2019towards}
Lambrecht, J., K{\"a}stner, L.: Towards the usage of synthetic data for marker-less pose estimation of articulated robots in rgb images. In: ICAR (2019)

\bibitem{lee2020icra:dream}
Lee, T.E., Tremblay, J., To, T., Cheng, J., Mosier, T., Kroemer, O., Fox, D., Birchfield, S.: Camera-to-robot pose estimation from a single image. In: International Conference on Robotics and Automation (ICRA) (2020)

\bibitem{lepetit2009epnp}
Lepetit, V., Moreno-Noguer, F., Fua, P.: Epnp: An accurate o (n) solution to the pnp problem. IJCV  (2009)

\bibitem{li2022self}
Li, S., De~Wagter, C., De~Croon, G.C.: Self-supervised monocular multi-robot relative localization with efficient deep neural networks. In: 2022 International Conference on Robotics and Automation (ICRA). pp. 9689--9695. IEEE (2022)

\bibitem{LI2023110491grasping}
Li, X., Zhang, X., Zhou, X., Chen, I.M.: Upg: 3d vision-based prediction framework for robotic grasping in multi-object scenes. Knowledge-Based Systems  \textbf{270},  110491 (2023)

\bibitem{lu2023markerless}
Lu, J., Richter, F., Yip, M.C.: Markerless camera-to-robot pose estimation via self-supervised sim-to-real transfer. In: Proceedings of the Conference on Computer Vision and Pattern Recognition (CVPR) (2023)

\bibitem{DexNet4}
Mahler, J., Matl, M., Satish, V., Danielczuk, M., DeRose, B., McKinley, S., Goldberg, K.: Learning ambidextrous robot grasping policies. Science Robotics  \textbf{4}(26),  eaau4984 (2019)

\bibitem{MASEHIAN2017188}
Masehian, E., Jannati, M., Hekmatfar, T.: Cooperative mapping of unknown environments by multiple heterogeneous mobile robots with limited sensing. Robotics and Autonomous Systems  \textbf{87},  188--218 (2017)

\bibitem{MING202114depth}
Ming, Y., Meng, X., Fan, C., Yu, H.: Deep learning for monocular depth estimation: A review. Neurocomputing  \textbf{438},  14--33 (2021)

\bibitem{Moon_2019_ICCV_3DMPPE}
Moon, G., Chang, J., Lee, K.M.: Camera distance-aware top-down approach for 3d multi-person pose estimation from a single rgb image. In: The IEEE Conference on International Conference on Computer Vision (ICCV) (2019)

\bibitem{Morrison_handineye}
Morrison, D., Corke, P., Leitner, J.: {Closing the Loop for Robotic Grasping: A Real-time, Generative Grasp Synthesis Approach}. In: Proc.\ of Robotics: Science and Systems (RSS) (2018)

\bibitem{olson2011apriltag}
Olson, E.: Apriltag: A robust and flexible visual fiducial system. In: ICRA (2011)

\bibitem{PAPADIMITRIOU2022117052collab}
Papadimitriou, A., Mansouri, S.S., Nikolakopoulos, G.: Range-aided ego-centric collaborative pose estimation for multiple robots. Expert Systems with Applications  \textbf{202},  117052 (2022)

\bibitem{Park_handtoeye}
Park, D., Seo, Y., Chun, S.Y.: Real-time, highly accurate robotic grasp detection using fully convolutional neural networks with high-resolution images (2019)

\bibitem{torchautograd}
Paszke, A., Gross, S., Chintala, S., Chanan, G., Yang, E., DeVito, Z., Lin, Z., Desmaison, A., Antiga, L., Lerer, A.: Automatic differentiation in pytorch  (2017)

\bibitem{7989190supervision}
Pinto, L., Davidson, J., Gupta, A.: Supervision via competition: Robot adversaries for learning tasks. In: 2017 IEEE International Conference on Robotics and Automation (ICRA). pp. 1601--1608 (2017)

\bibitem{prusak2008pose}
Prusak, A., Melnychuk, O., Roth, H., Schiller, I., Koch, R.: Pose estimation and map building with a time-of-flight-camera for robot navigation. International Journal of Intelligent Systems Technologies and Applications  \textbf{5}(3-4),  355--364 (2008)

\bibitem{9988012ICECCME}
Rana, A., Vulpi, F., Galati, R., Milella, A., Petitti, A.: A pose estimation algorithm for agricultural mobile robots using an rgb-d camera. In: 2022 International Conference on Electrical, Computer, Communications and Mechatronics Engineering (ICECCME). pp.~1--5 (2022)

\bibitem{Rizk2019CooperativeHM}
Rizk, Y., Awad, M., Tunstel, E.W.: Cooperative heterogeneous multi-robot systems. ACM Computing Surveys (CSUR)  \textbf{52},  1 -- 31 (2019)

\bibitem{Salvini2021SafetyCE}
Salvini, P., Paez-Granados, D.F., Billard, A.: Safety concerns emerging from robots navigating in crowded pedestrian areas. International Journal of Social Robotics  \textbf{14},  441 -- 462 (2021)

\bibitem{sun2019deep}
Sun, K., Xiao, B., Liu, D., Wang, J.: Deep high-resolution representation learning for human pose estimation. In: CVPR (2019)

\bibitem{integral}
Sun, X., Xiao, B., Wei, F., Liang, S., Wei, Y.: Integral human pose regression. In: Proceedings of the European Conference on Computer Vision (ECCV) (2018)

\bibitem{5152690}
Svenstrup, M., Tranberg, S., Andersen, H.J., Bak, T.: Pose estimation and adaptive robot behaviour for human-robot interaction. In: 2009 IEEE International Conference on Robotics and Automation. pp. 3571--3576 (2009)

\bibitem{tian2023robot}
Tian, Y., Zhang, J., Yin, Z., Dong, H.: Robot structure prior guided temporal attention for camera-to-robot pose estimation from image sequence. In: Proceedings of the Conference on Computer Vision and Pattern Recognition (CVPR) (2023)

\bibitem{TZAFESTAS2014635}
Tzafestas, S.G.: 15 - mobile robots at work. In: Tzafestas, S.G. (ed.) Introduction to Mobile Robot Control, pp. 635--663. Elsevier, Oxford (2014)

\bibitem{uchibe1996vision}
Uchibe, E., Asada, M., Noda, S., Takahashi, Y., Hosoda, K.: Vision-based reinforcement learning for robocup: Towards real robot competition. In: Proc. of IROS. vol.~96 (1996)

\bibitem{privacy-preserving}
Xia, Y., Tang, Y., Hu, Y., Hoffman, G.: Privacy-preserving pose estimation for human-robot interaction. CoRR  \textbf{abs/2011.07387} (2020)

\bibitem{9255211humanrobot}
Xu, C., Yu, X., Wang, Z., Ou, L.: Multi-view human pose estimation in human-robot interaction. In: IECON 2020 The 46th Annual Conference of the IEEE Industrial Electronics Society. pp. 4769--4775 (2020)

\bibitem{xun2023crepes}
Xun, Z., Huang, J., Li, Z., Xu, C., Gao, F., Cao, Y.: Crepes: Cooperative relative pose estimation towards real-world multi-robot systems. arXiv preprint arXiv:2302.01036  (2023)

\bibitem{yang2021reactive}
Yang, W., Paxton, C., Mousavian, A., Chao, Y.W., Cakmak, M., Fox, D.: Reactive human-to-robot handovers of arbitrary objects. In: IEEE International Conference on Robotics and Automation (ICRA). IEEE (2021)

\bibitem{zhou2019continuity}
Zhou, Y., Barnes, C., Lu, J., Yang, J., Li, H.: On the continuity of rotation representations in neural networks. In: Proceedings of the IEEE/CVF Conference on Computer Vision and Pattern Recognition. pp. 5745--5753 (2019)

\bibitem{zhu2023human}
Zhu, W., Ma, X., Ro, D., Ci, H., Zhang, J., Shi, J., Gao, F., Tian, Q., Wang, Y.: Human motion generation: A survey. IEEE Transactions on Pattern Analysis and Machine Intelligence  (2023)

\bibitem{zuo2019craves}
Zuo, Y., Qiu, W., Xie, L., Zhong, F., Wang, Y., Yuille, A.L.: Craves: Controlling robotic arm with a vision-based economic system. In: CVPR (2019)

\end{thebibliography}

\clearpage
\section*{Appendix}
The content of our supplementary material is organized as follows.

1. Implementation details of our method.

2. Additional quantitative results.

3. Additional qualitative results.

4. Error analysis of each sub-module acorss training stages.

5. Multi-view fusion results.

6. Failure cases of our method.

\section*{S1. Implementation Details}
\label{sec:implementation_details}

\subsection*{S1.1 Model architecture}
\label{sec:model_architecture}
We implement the proposed pipeline using PyTorch~\cite{torchautograd}. We utilize HRnet-w32~\cite{sun2019deep} as the feature extractor for \dn. \jn, \rn and \kn share a ResNet-50~\cite{he2016deep} network as the feature extractor.

\subsection*{S1.2 Method Details}
\label{sec:method_details}
We provide a detailed illustration of the relations of the variables mentioned in~\cref{sub:Overview}:
\begin{equation}
    \mathbf{P}^{'}_i = \mathbf{P}^{r}_i + (0, 0, d),   i = 0, 1, 2, ...
    \label{eq:relation_1}
\end{equation}
\begin{equation}
    \mathbf{t} = \mathbf{P}^{'}_{RootId}
    \label{eq:relation_2}
\end{equation}
\begin{equation}
    \mathbf{P} = FK(\mathbf{q}, \mathbf{R}, \mathbf{t})
    \label{eq:relation_3}
\end{equation}
where $d$, $RootId$, $\mathbf{q}$, $\mathbf{R}$, $\mathbf{t}$, $\mathbf{P}^{r}$, $\mathbf{P}^{'}$, and $\mathbf{P}$ are the depth (z-axis value) of the root keypoint, index of the root keypoint, joint states, camera-to-robot rotation, camera-to-robot translation, root-relative keypoint locations, \kn-derived keypoint locations and FK-based keypoint locations, respectively. $FK$ denotes the forward-kinematics process. 

Here we clarify the coordinate frames of the three sets of keypoints: $\mathbf{P}^{r}$ is a set of root-relative keypoints in the root-relative frame, with absolute xy value and root-relative z-axis offsets. Both $\mathbf{P}^{'}$ and $\mathbf{P}$ are absolute keypoints in the camera frame. $\mathbf{P}^{'}$ is derived from $d$ and $\mathbf{P}^{r}$ using~\cref{eq:relation_1}, and $\mathbf{P}$ is derived from forward kinematics as in~\cref{eq:relation_3}.

\subsection*{S1.3 Robot arms}
\label{sec:robot_arms}
We leverage the datasets provided by DREAM~\cite{lee2020icra:dream}, which contains images for three types of robot arms, namely Franka Emika Panda (Panda), Kuka iiwa7 (Kuka), and Rethink Robotics Baxter (Baxter).

Panda features 8 one-DoF joints and 7 keypoints, manually defined at joint articulations. The first 7 joints in Panda are revolute, facilitating pure rotational motion around a shared axis, and their internal state measures in degrees. The last joint is prismatic, allowing linear sliding along a shared axis without rotational motion, with its internal state measurable in millimeters.

Kuka comprises 7 one-DoF revolute joints and includes 8 defined keypoints. Baxter incorporates 15 one-DoF revolute joints and 17 defined keypoints. Both Kuka and Baxter have no prismatic joints.

Panda and Kuka are both single-arm robotic manipulators, while Baxter has multiple arms connected with a torso, resulting in a human-like appearance.

\subsection*{S1.4 Rendering method}
We utilize the provided Unified Robot Description Format (URDF) file following previous works \cite{labbe2021robopose, lu2023markerless}.
With URDF parsed and joint configuration \textbf{q} given, we determine the robot's morphology via forward kinematics, using the URDFPytorch module provided by \cite{labbe2021robopose}. \textbf{R} and \textbf{t} determine the cam-to-robot-root transformation. As the robot's own morphology and global pose are acquired, we render its mask with a given camera using a mesh renderer derived from \cite{lu2023markerless}.

\section*{S2. Additional Quantitative Results}
\label{sec:additional_quantitative}
\begin{table}[htbp]
    \centering
    \begin{minipage}[t]{0.47\textwidth}
        \centering
        \caption{Ablation studies on the influence of \kn on \jn.}
        \label{table:ablation_keypointnet_on_jointnet}
        \small
        \setlength{\tabcolsep}{1.2mm}
        \resizebox{\linewidth}{!}{
        \begin{tabular}{lccc}
        \toprule
        Dataset & 
        KeypointNet & 
        \multicolumn{2}{c}{Mean Joint Error $\color{red}\downarrow$}  \\
        \midrule
        Panda DR & $-$ & 14.49$^\circ$ & 3.45 mm    \\
        Panda DR & $+$ & \textbf{9.29}$^\circ$ &  \textbf{2.63} mm \\
        \midrule
        Panda Photo & $-$ & 15.14$^\circ$ & 3.83 mm  \\
        Panda Photo & $+$ & \textbf{10.45}$^\circ$ & \textbf{2.69} mm \\
        \midrule
        Kuka DR & $-$ & \multicolumn{2}{c}{9.59$^\circ$}  \\
        Kuka DR & $+$ & \multicolumn{2}{c}{\textbf{8.25}$^\circ$}  \\
        \midrule
        Kuka Photo & $-$ & \multicolumn{2}{c}{9.51$^\circ$}  \\
        Kuka Photo & $+$ & \multicolumn{2}{c}{\textbf{8.22}$^\circ$}  \\
        \midrule
        Baxter DR & $-$ & \multicolumn{2}{c}{13.73$^\circ$}  \\
        Baxter DR & $+$ & \multicolumn{2}{c}{\textbf{12.95}$^\circ$}  \\
        \bottomrule
        \end{tabular}}  
    \end{minipage}
    \hfill
    \begin{minipage}[t]{0.47\textwidth}
        \centering
        \caption{Ablation studies on the influence of \kn on \rn.}
        \label{table:ablation_keypointnet_on_rotationnet}
        \small
        \setlength{\tabcolsep}{0.6mm}
        \resizebox{\linewidth}{!}{
        \begin{tabular}{lcc}
        \toprule
        Dataset & KeypointNet & Mean Rot. Error (${}^\circ$) $\color{red}\downarrow$ \\
        \midrule
        Panda DR & $-$ & 3.78 \\
        Panda DR & $+$ & $\textbf{3.25}$  \\
        \midrule
        Panda Photo & $-$ & 4.03 \\
        Panda Photo & $+$ & $\textbf{3.36}$ \\
        \midrule
        Kuka DR & $-$ & 8.14 \\
        Kuka DR & $+$ & $\textbf{7.12}$ \\
        \midrule
        Kuka Photo & $-$ & 8.28 \\
        Kuka Photo & $+$ & $\textbf{7.38}$ \\
        \midrule
        Baxter DR & $-$ & 4.36 \\
        Baxter DR & $+$ & $\textbf{3.69}$ \\
        \bottomrule
        \end{tabular}}
    \end{minipage}
\end{table}

\subsection*{S2.1 Ablation results on the influence of \kn on \jn and \rn}
\label{sec:ablation_kn}
From the results in~\cref{table:ablation_keypointnet_on_jointnet,table:ablation_keypointnet_on_rotationnet}, we can infer that the presence of \kn significantly improves the prediction accuracy of \jn and \rn. This strongly validates our model design. By sharing a feature extractor with \jn and \rn, \kn guides the multi-module network towards learning pixel-aligned visual features. This bridges the gap between the pixel space and the robotic parameter space, providing strong support for the holistic robot pose estimation task.

\begin{table}[ht]
    \footnotesize
    \centering
    \caption{Performance comparison with different numbers of in-frame keypoints (different levels of truncation severity).}
    \label{table:ablation_inframe_kp}
    \setlength{\tabcolsep}{3pt}
    \resizebox{0.75\linewidth}{!}{ 
    \begin{tabular}{l|c|c|cc|cc}
    \toprule
    \multirow{2}{*}{Dataset} &
    In-frame & 
    \multirow{2}{*}{Images} & 
    \multicolumn{2}{c|}{RoboPose~\cite{labbe2021robopose}}&  
    \multicolumn{2}{c}{Ours}
    \\
    & KPs & & AUC $\color{green}\uparrow$ & Mean $\color{red}\downarrow$ & AUC $\color{green}\uparrow$ & Mean $\color{red}\downarrow$\\
    \midrule
    Panda 3CAM-AK&4 & 96 & 52.28 & 47.71 & \textbf{62.54} & \textbf{37.45} \\ 
    Panda 3CAM-AK&5 & 92 & 73.88 & 26.11 & \textbf{79.31} & \textbf{20.68} \\ 
    Panda 3CAM-AK&6 & 105 & 6.05 & 126.87 & \textbf{71.98} & \textbf{28.01} \\
    Panda 3CAM-AK&7 & 6041 & 72.32 & 31.58 & \textbf{82.88} & \textbf{18.29} \\ 
    \midrule
    Panda 3CAM-XK&7 &4966 &\textbf{77.61} &\textbf{22.27} & 76.01 & 23.99  \\
    \midrule
    Panda 3CAM-RS&5 & 100 & 38.60 & 61.39 & \textbf{76.57} & \textbf{23.42} \\ 
    Panda 3CAM-RS&6 & 11 & 35.83 & 64.16 & \textbf{74.56} & \textbf{25.42} \\ 
    Panda 3CAM-RS&7 & 5833 & 74.90 & 25.42 & \textbf{75.20} & \textbf{24.79} \\ 
    \midrule
    Panda ORB&4 & 98 & \textbf{16.96} & 88.73 & 14.40 & \textbf{85.66} \\ 
    Panda ORB&5 & 476 & 47.75 & 52.42 & \textbf{52.02} & \textbf{47.97} \\ 
    Panda ORB&6 & 650 & 56.22 & 44.50 & \textbf{67.78} & \textbf{32.21} \\
    Panda ORB&7 & 31091 & 71.20 & 29.28 & \textbf{75.94} &\textbf{24.05} \\
    \midrule
    Panda Photo&2 & 4 & 36.05 & 186.04 & \textbf{38.57} & \textbf{69.22} \\ 
    Panda Photo&3 & 17 & \textbf{53.44} & 112.83 & 38.13 & \textbf{75.10} \\
    Panda Photo&4 & 491 &\textbf{66.65} & 49.85 & 66.52 &\textbf{34.80} \\ 
    Panda Photo&5 & 207 & 74.06 & 36.89 & \textbf{76.74} & \textbf{23.57} \\
    Panda Photo&6 & 202 & 73.80 & 43.72 &\textbf{78.43} &\textbf{21.82} \\ 
    Panda Photo&7 & 5076 & \textbf{84.12} & 24.76 & 84.01 & \textbf{16.21} \\ 
    \midrule
    Panda DR & 2 & 4 & 0 &17347.83 & \textbf{23.00}&\textbf{77.40}\\
    Panda DR & 3 & 10 & 0 &15646.74 & \textbf{61.24} &\textbf{38.75 }\\
    Panda DR & 4 &391 & 52.19 & 58.09 &\textbf{ 68.04} &\textbf{33.56} \\
    Panda DR & 5 &231 & 71.76 & 33.37 & \textbf{78.21} &\textbf{21.83 }\\
    Panda DR & 6 &206 & 78.16 & 25.66 & \textbf{79.59} &\textbf{22.30} \\
    Panda DR & 7 & 5156 & \textbf{86.02} & 18.40 & 84.20 & \textbf{16.44}\\
    \bottomrule
    \end{tabular}}
\end{table}

\subsection*{S2.2 Ablation results with varying levels of truncation severity}

We observe that image truncation, \ie, the partial capture of robot arms in images, considerably influences the estimation results. To evaluate our method's robustness, we measure our model performance with varying numbers of in-frame keypoints (representing different levels of truncation severity) and compare the results with RoboPose~\cite{labbe2021robopose} in the context of holistic estimation with unknown joint states, as detailed in~\cref{table:ablation_inframe_kp}.
Our approach generally surpasses RoboPose~\cite{labbe2021robopose} across various datasets and truncation degrees, particularly in terms of mean estimation errors. This demonstrates the improved robustness of our approach.

\subsection*{S2.3 Ablation results with different selection of the reference root keypoint}
In our experiments for all three robots, we consistently designate the keypoint located in close proximity to the geometric center of the robot as the reference root keypoint. This choice is motivated by the observation that the keypoints near the geometric center are less likely to be truncated. 

\begin{table}[h]
    \footnotesize
    \centering
    \caption{Ablation studies on the selection of reference root keypoint on the Panda DR dataset. }
    \label{table:ablation_reference_root}
    \setlength{\tabcolsep}{1mm}
    \resizebox{0.55\linewidth}{!}{
    \begin{tabular}{lc}
    \toprule
    Reference Root Keypoint & Mean Depth Error ($mm$) $\color{red}\downarrow$  \\
    \midrule
    $\mathbf{P}_{0}$ & 18.07  \\
    $\mathbf{P}_{1}$ & 14.58 \\
    $\mathbf{P}_{2}$ & 13.00  \\
    $\mathbf{P}_{3}$ & \textbf{9.45}  \\
    $\mathbf{P}_{4}$ & 15.57  \\
    $\mathbf{P}_{5}$ & 18.13   \\
    $\mathbf{P}_{6}$ & 20.38   \\
    \bottomrule
    \end{tabular}}
\end{table}

\begin{figure}[h]
    \centering
    \includegraphics[width=0.4\linewidth]{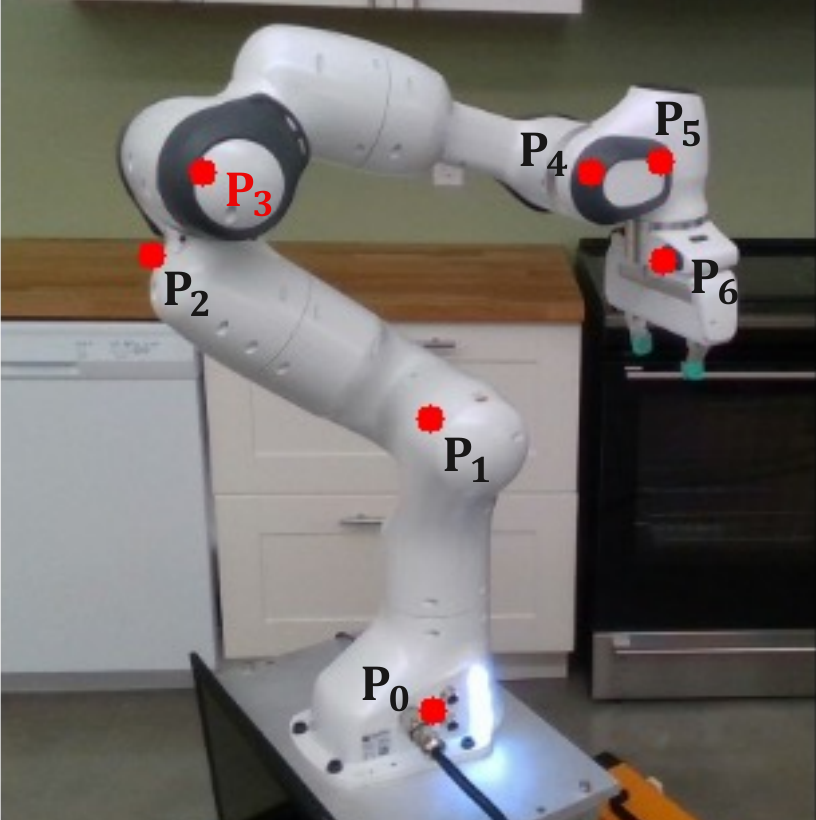}
    \caption{Visualization of the keypoints of robot Panda. We choose $\mathbf{P}_{3}$, the keypoint closest to the geometric center, as the reference root keypoint.}
    \label{fig:keypoint}
\end{figure}

In addition, we quantitatively evaluate the selection of the reference root keypoint on the Panda DR dataset. We measure Mean Depth Error, which represents the mean error of the root depth predictions of \dn, under different reference root keypoints, denoted as $\mathbf{P}_{i}  (i \in {0,1,2,3,4,5,6})$. \Cref{table:ablation_reference_root} and \Cref{fig:keypoint} demonstrate that the central keypoint indeed leads to more precise depth estimations, validating our choices. 

\subsection*{S2.4 Ablation results with different model backbones}
In our SOTA model, we use HRNet-32 for \dn (scale-variant) and ResNet-50 for other modules (scale-invariant). Here we present a comparison of other options in~\cref{table:ablation_backbone}, which validates our selection. Lighter backbones generally yield slightly inferior results, though they are still comparable and may be suitable in scenarios where speed is prioritized.

\begin{table}[htbp]
    \vspace{-0.1cm}
    \centering
    \small
    \caption{Ablation studies on the choices of model backbone. }
    \label{table:ablation_backbone}
    \setlength{\tabcolsep}{1mm}
    \resizebox{0.75\linewidth}{!}{
    \begin{tabular}{cccccc}
    \toprule
     \multirow{2}{*}{\dn}  & \multirow{2}{*}{Other modules}  & \multicolumn{2}{c}{Panda DR} & \multicolumn{2}{c}{Panda Photo}  \\
     &  & AUC $\color{green}\uparrow$ & Mean (mm)$\color{red}\downarrow$ & AUC $\color{green}\uparrow$ & Mean (mm)$\color{red}\downarrow$ \\
    \midrule %
     ResNet-34 & ResNet-50 & 80.7 & 20.0 & 80.5 & 19.9 \\
     ResNet-50 & ResNet-50 & 79.5 & 21.2 & 78.8 & 21.7 \\ 
     HRNet-32 & ResNet-34 & 78.3 & 22.5 & 78.0 & 22.5 \\
     \rowcolor{mygray}
     HRNet-32 & ResNet-50 & \textbf{82.7} & \textbf{18.0} & \textbf{82.0} & \textbf{18.4} \\ 
     HRNet-32 & HRNet-32 & 79.9 & 20.9 & 79.0 & 21.5 \\ 
    \bottomrule
    \end{tabular}}
    \vspace{-0.5cm}
\end{table}

\section*{S3. Additional Qualitative Results}
\subsection*{S3.1 Results with unknown joint states}
\begin{figure*}[t]
    \centering
   \includegraphics[width=0.85\linewidth]{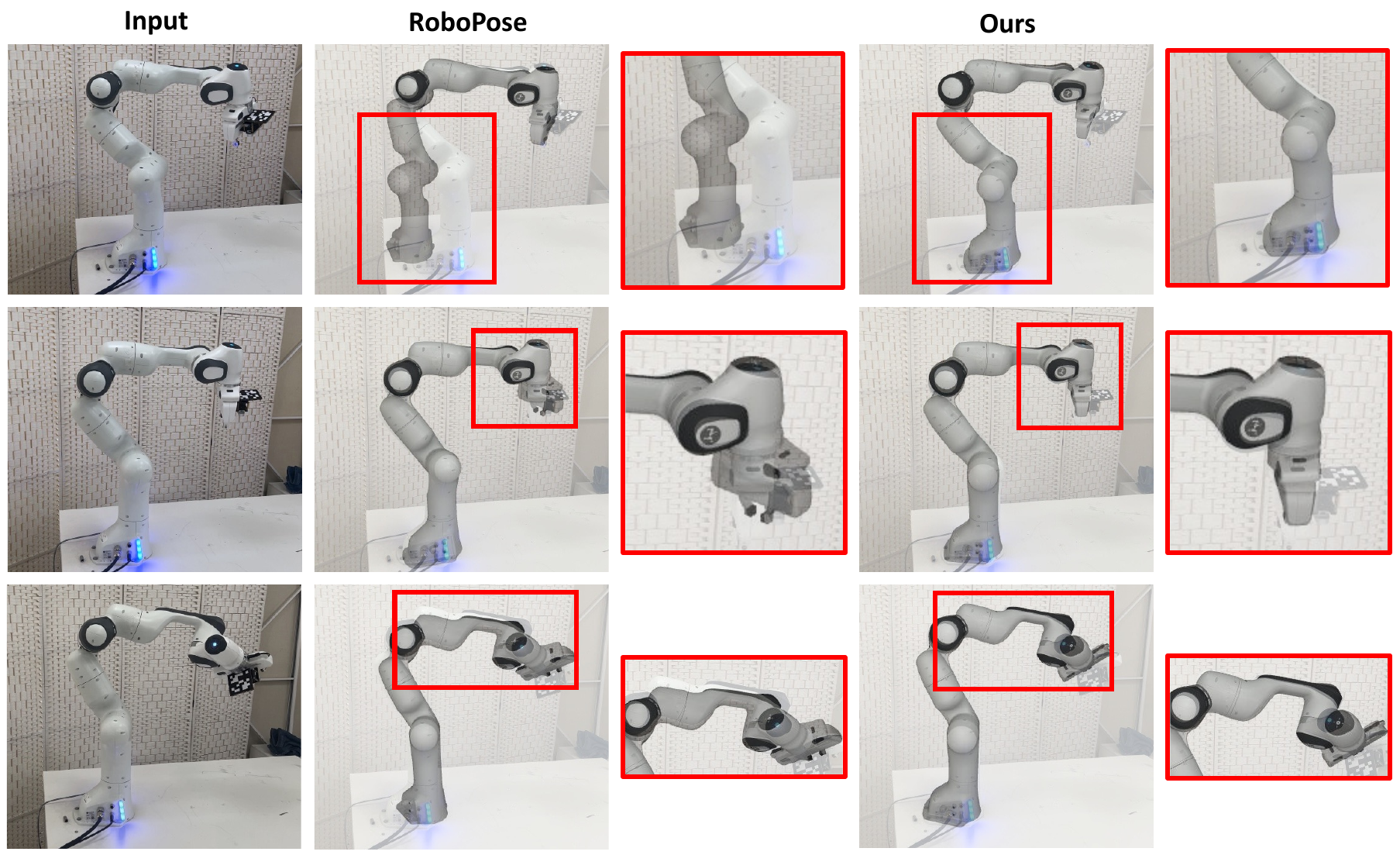}
   \caption{Additional qualitative comparison results on in-the-wild images, with unknown joint states.}
   \label{fig:qualitative_real}
\end{figure*}

Additionally, in the context of holistic estimation with unknown joint states, we present qualitative comparison results on freshly-collected in-the-wild lab images, demonstrating our method's real-world generalization capabilities.
The images are captured using an iPhone, and camera calibration is not performed. Solely based on RGB images of the robot Panda and predefined camera intrinsic parameters, we estimate both the robot's pose (6DoF) and joint states (8DoF) using our approach. The estimated robot pose and states are rendered and visualized in~\cref{fig:qualitative_real}. The marker mounted at the end-effector is not utilized.

For example, in the top row of images, our method provides a more accurate estimation of the pose of the robot's base while RoboPose~\cite{labbe2021robopose} fails to properly position the base. In the bottom row of images, our method provides a more precise estimation of the joint angles of the robot's upper part, resulting a more accurate alignment between the rendered robot image and the input robot image.

The key component that contributes to our method's generalization capabilities is the self-supervision of our framework. It enables our model to be fine-tuned on unlabeled data, making our method well-suited for adaptation to various real-world use cases. To further investigate this, we also conducted self-supervised training with 50\% real data from the correspondent DREAM~\cite{lee2020icra:dream} subset, resulting in only a slight decrease in AUC: \underline{81.9} \vs 82.2 on Panda 3CAM-AK and \underline{72.6} \vs 75.2 on Panda 3CAM-RS. This demonstrates that self-supervised domain adaptation can be effectively applied even with a small amount of data, and sets the foundation for more flexible real-world applications of our method.

\subsection*{S3.2 Results with known joint states}
\begin{figure*}[t]
    \centering
   \includegraphics[width=0.78\linewidth]{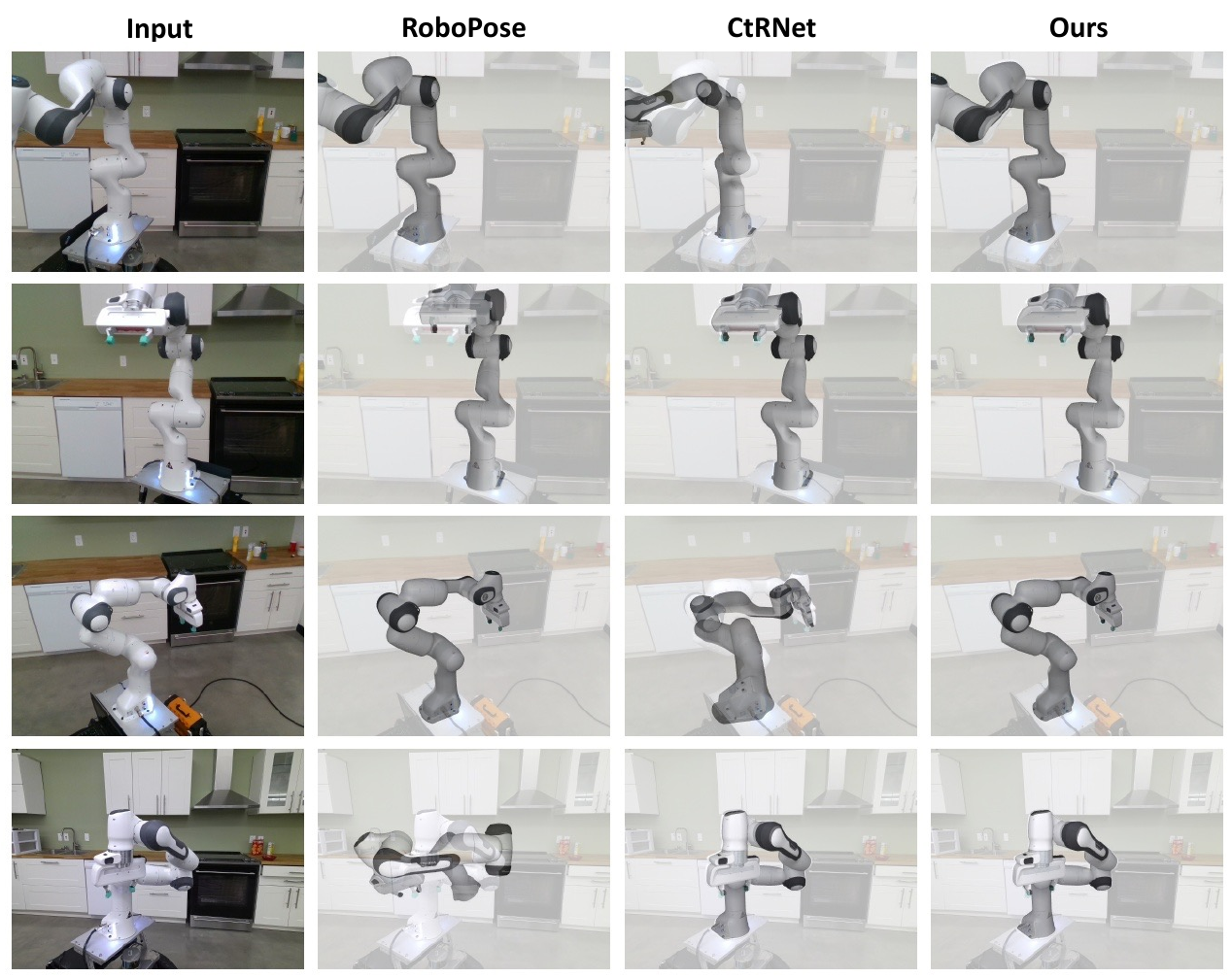}
   \caption{Additional qualitative comparison results on real-world images, with known joint states.}
   \label{fig:qualitative_known}
\end{figure*}

In the context of estimation with known joint states, we also present qualitative comparison results on several real world images of robot Panda from the DREAM~\cite{lee2020icra:dream} datasets, as shown in~\cref{fig:qualitative_known}.
Our approach can be easily adapted for known-joint estimation by simply substituting predicted joint states with ground-truth joint states. 

Moreover, by eliminating the inefficient Render-and-Compare (RnC) optimization used RoboPose~\cite{labbe2021robopose} and unstable Perspective-n-Points (PnP) methods used in CtRNet~\cite{lu2023markerless}, our approach generally outperforms previous state-of-the-art methods in this well-explored setting.
Free of the cumbersome RnC, our approach achieves a 12$\times$ speed boost compared with RoboPose~\cite{labbe2021robopose}. Without unstable and error-prone PnP methods, our approach demonstrates higher estimation accuracy and robustness compared with CtRNet~\cite{lu2023markerless}.
Please kindly refer to the video on the project page for more video comparison results.

\section*{S4. Error Analysis}
 We further provide error analysis of each module across training stages (training on synthetic data and self-supervised training on real data) on Panda 3CAM-AK in~\cref{table:error_analysis}. 
 Errors of the 4 modules (\jn, \rn, \kn and \dn) after each training stage are presented. The results show that our multi-stage training effectively reduces the errors, rather than accumulating them. 
 Moreover, our end-to-end training could mitigate error propagation by jointly training all modules, allowing supervision signals to guide inter-module learning and prediction refinement. This contrasts with traditional RnC and non-differentiable PnP systems \cite{labbe2021robopose,lee2020icra:dream}, where error recovery is limited by the lack of integrated training. These empirical results clearly demonstrate the convergence capability of our approach.

\begin{table}[t]
    \center
    \small
    \caption{Error analysis of each module across training stages on Panda 3CAM-AK dataset.}
    \label{table:error_analysis}
    \setlength{\tabcolsep}{2mm}
    \resizebox{\linewidth}{!}{ 
    \begin{tabular}{l l| l l | l}
    \thickhline
    Module  & Metric  & Synthetic &  Self-supervised & Err. Change \\
    \hline
    JointNet  &  $\mathcal{L}_1$ Joint Angle Err. ($^\circ$)      &   9.68     & 8.83 & $\downarrow$8.8\% \\
    RotationNet  &  $\mathcal{L}_1$ Euler Rotation Angle Err. ($^\circ$)   &   2.94   &   2.01 & $\downarrow$31.6\% \\
    KeypointNet &  Keypoint Euclidean Dis. (pixel)   &    2.38      &  1.97 & $\downarrow$17.2\% \\
    DepthNet  &  $\mathcal{L}_1$ Root Depth Err. (mm)      &   14.1     &  11.8 & $\downarrow$16.3\% \\
    \thickhline
    \end{tabular}} 
\end{table}

\section*{S5. Multi-view Fusion Results}
Although our primary focus is on monocular estimation to ensure a fair comparison with previous studies, we also extend our method to a multi-view setting through quantitative prediction fusion on the Panda-Orb subset, which contains RGB sequences of the movement of a Panda robot arm from 27 viewpoints. 
We ensemble the predictions from the first view with each of the remaining 26 views. Specifically, for each of the remaining 26 views whose predictions are being fused, we average the joint state predictions, transform the camera-to-robot translation predictions of the first view into the current viewpoint using relative camera poses, and average the pose-aligned translation predictions. For simplicity, we do not fuse the camera-to-robot rotation predictions, because the process of averaging two $SO(3)$ rotations has no analytical solution.

This fusion approach results in a 2.8\% average increase in AUC$\color{green}\uparrow$ and a 9.1\% decrease in mean ADD$\color{red}\downarrow$ for the remaining 26 views, demonstrating the usefulness of multi-view information in promoting robustness. Future research could further investigate the integration of multi-view information into the framework and model design.

\section*{S6. Failure Cases}
\begin{figure}[t]
\centering
   \includegraphics[width=0.7\linewidth]{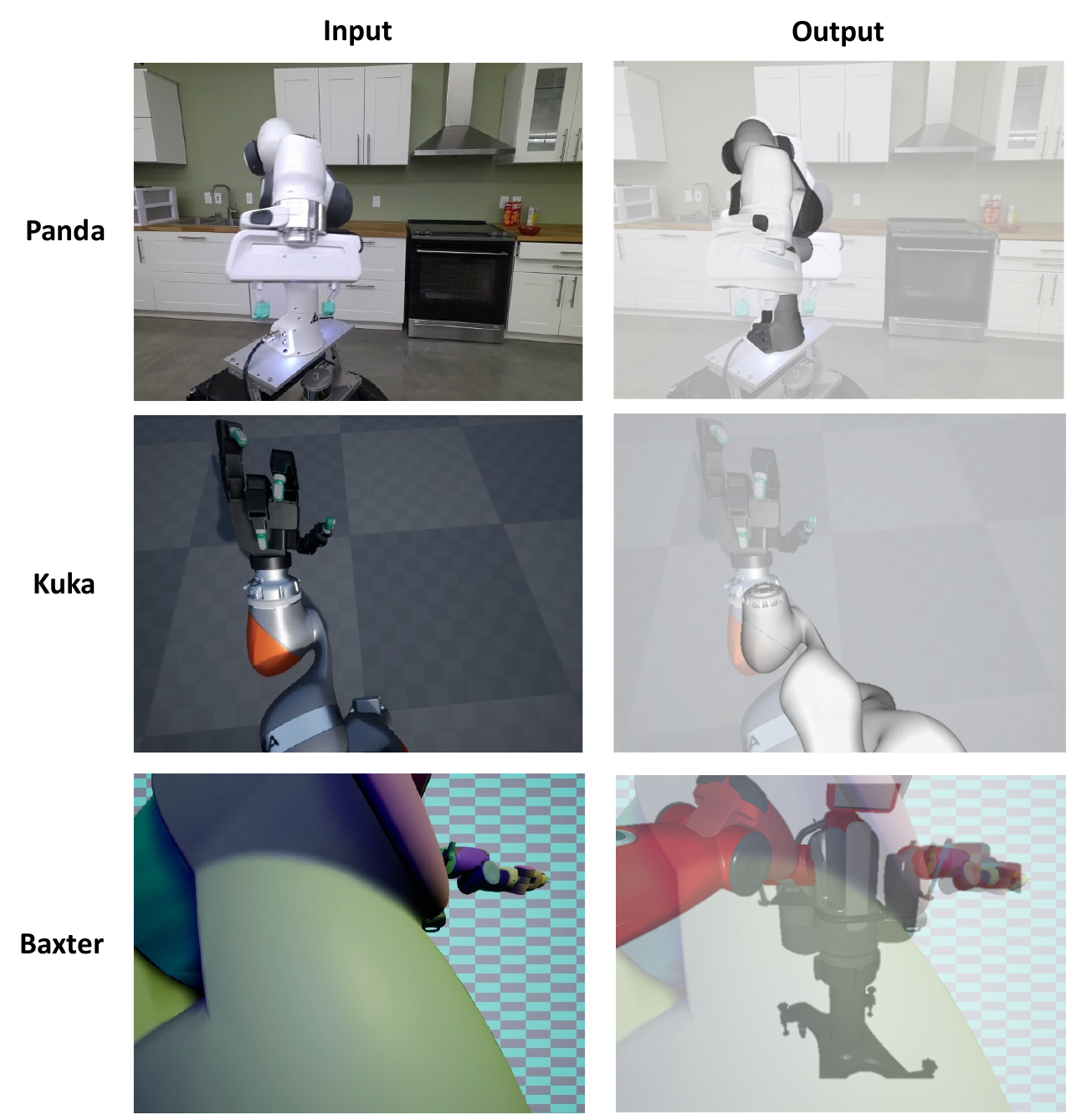}
   \caption{Typical failure cases.}
   \label{fig:failure}
\end{figure}

We illustrate the typical failure cases of our method for each robot arm in~\cref{fig:failure}. In the case of Panda, estimation becomes unstable under extremely severe self-occlusion. For the example of Kuka and Baxter, estimation fails due to an extremely severe level of truncation. When only a very small number of keypoints are visible, the estimations are generally more error-prone, as also evidenced in~\cref{table:ablation_inframe_kp}.

\end{document}